\documentclass[10pt,twocolumn,letterpaper]{article}

\usepackage[pagenumbers]{wacv} 

\usepackage{graphicx}
\usepackage{amsmath}
\usepackage{amssymb}
\usepackage{booktabs}
\usepackage{multirow}
\usepackage{adjustbox}
\usepackage{subcaption}
\usepackage{tabularx}

\usepackage{hyperref}
\hypersetup{pagebackref,breaklinks,colorlinks}

\def\wacvPaperID{394} 
\def\confName{WACV}
\def\confYear{2025}

\begin{document}

\title{Online-LoRA: Task-free Online Continual Learning via Low Rank Adaptation}

\author{Xiwen Wei\\
\small{The University of Texas at Austin}\\
{\tt\small xiwenwei@utexas.edu}
\and
Guihong Li \\
\small{AMD}\\
{\tt\small liguihong1995@gmail.com}
\and
Radu Marculescu \\
\small{The University of Texas at Austin}\\
{\tt\small radum@utexas.edu}
}

\maketitle

\begin{abstract}
     Catastrophic forgetting is a significant challenge in online continual learning (OCL), especially for non-stationary data streams that do not have well-defined task boundaries. This challenge is exacerbated by the memory constraints and privacy concerns inherent in rehearsal buffers. To tackle catastrophic forgetting, in this paper, we introduce \textit{Online-LoRA}, a novel framework for task-free OCL. Online-LoRA allows to finetune pre-trained Vision Transformer (ViT) models in real-time to address the limitations of rehearsal buffers and leverage pre-trained models’ performance benefits. As the main contribution, our approach features a novel online weight regularization strategy to identify and consolidate important model parameters. Moreover, Online-LoRA leverages the training dynamics of loss values to enable the automatic recognition of the data distribution shifts. Extensive experiments across many task-free OCL scenarios and benchmark datasets (including CIFAR-100, ImageNet-R, ImageNet-S, CUB-200 and CORe50) demonstrate that Online-LoRA can be robustly adapted to various ViT architectures, while achieving better performance compared to SOTA methods \footnote{Code: \href{https://github.com/Christina200/Online-LoRA-official.git}{https://github.com/Christina200/Online-LoRA-official.git}}. 
\end{abstract}

\vspace{-20pt}
\section{Introduction}
\label{sec:intro}

Continual learning (CL) is pivotal in enabling machine learning systems to learn new concepts while preserving the previously learned knowledge. This ability is crucial for real-time applications like robotics, healthcare, and autonomous driving \cite{app-1, app-2}. However, a major hurdle in CL is \textit{catastrophic forgetting}, where learning new information impairs performance on previously learned data. 

Existing CL methods are typically classified along two dimensions: (1) task-based or task-free, depending on whether the boundaries between different tasks are known \cite{task-free-1, task-free-2, task-free-3}; and (2) online or offline, based on whether the setting allows for multiple iterations over the data (offline) or requires a single pass through the data (online) \cite{ocl-1, ocl-2, ocl-3}. 

In offline task-based CL, a series of tasks is presented sequentially. Typically, it is assumed that each task comprises a dataset with samples drawn from a distinct independent and identically distributed (i.i.d.) distribution \cite{iid-setting, iid-setting2}. The samples are i.i.d. within the same task, though across tasks the distributions may differ. Additionally, it is assumed that the probability distribution from which the data is drawn remains stationary between the training and inference phases for any given task \cite{SI, gem}. This assumption simplifies the learning problem by ensuring that the trained model can be effectively applied to new data from the same task without the need to account for distributional shifts \cite{mas_ocl}. 

While offline task-based CL has paved the way for understanding how models can sequentially learn, its assumptions are often misaligned with the intricacies of real-world data. For instance, the need of completed training before inference, known as the lack of anytime inference capability, does not hold for applications where decisions must be made on-the-fly based on data that is continuously changing. Moreover, the offline task-based CL assumes well-defined task boundaries, a condition seldom met outside tightly controlled experimental environments. In contrast, real-world data streams are inherently continuous and lack clear task boundaries, often exhibiting gradual transitions. 

Motivated by the limitations of offline task-based setting, in this paper we focus on \textit{task-free online CL}; this scenario is characterized by the constraints of seeing a stream of samples only once, and the absence of knowledge regarding task identities and task boundaries during both training and inference \cite{cn-dpm, cope, dyson, dcm}. 

Pre-trained Vision Transformers \cite{vit} have demonstrated superior performance on various vision tasks, hence integrating them into CL has attracted increasing interest \cite{slca, l2p, unified-cl-peft}. Indeed, the extensive prior knowledge of pre-trained models enhances knowledge transfer \cite{cl-survey}, brings significant performance improvements compared to traditional SOTA methods trained from scratch, and provides robust generalizability and adaptability, especially valuable in data-scarce environments \cite{why_pretrain, pretrained-all-you-need}. Recent studies \cite{task-arithmetic, color, coda-prompt} have demonstrated the potential of using parameter efficient fine-tuning (PEFT) techniques like prompt tuning \cite{prompt_tuning} and Low-Rank Adaptation (LoRA) \cite{hu2021lora} with pre-trained models for offline task-based CL. 


Given the need for task-free OCL and the advantages provided by pre-trained models and PEFT methods, we wonder, \textit{whether task-free OCL can benefit from pre-trained models and PEFT as effectively as offline task-based CL}. To this end, we propose \textbf{Online-LoRA}, a new approach integrating pre-trained ViT and LoRA into the task-free OCL scenario. Online-LoRA learns incrementally with each new piece of information. More precisely, we propose an extensible architecture that expands the model with additional LoRA parameters where the loss surface plateaus \cite{mas_ocl}. Thus, by utilizing loss plateaus to recognize shifts in data distribution, our model remains robust in continuously changing environments. Furthermore, we propose a new online parameter regularization, aimed to mitigate forgetting and enhance memory efficiency. In our regularization, the importance weights are calculated on the LoRA parameters exclusively, rather than using the entire set of model parameters like in EWC \cite{ewc}. This decreases the computational and memory requirements significantly, thus enabling online updates of the model parameter importance throughout the learning process.  

We summarize our main contributions as follows: 
\begin{itemize}
    \item  We propose Online-LoRA, an innovative approach that can efficiently learn from changing data in an online, task-free manner, thus enabling inference at any time. We achieve this through continual low rank adaptation and automatic detection of data distribution shifts based on loss plateaus. 
    \item We present an online weight regularization mechanism that effectively mitigates forgetting by adapting the estimation of model parameter importance to the incoming data with minimal additional memory. We achieve this by using a Laplace approximation to estimate the uncertainty around the LoRA parameters. 
    \item Our extensive evaluations with various ViT architectures across multiple task-free OCL benchmarks, under the settings of class and domain incremental learning, demonstrate that Online-LoRA consistently outperforms existing SOTA methods. Moreover, Online-LoRA exhibits robust performance across various task sequence lengths and ViT architectures, showcasing its effectiveness in diverse learning contexts. 
\end{itemize}


\section{Related work}
\label{related-work}

\subsection{Continual learning}

Since many existing CL methods are offline task-based, their transition to the online, task-free setting is not trivial. Here, we discuss four categories of CL methods and their adaptability to task-free OCL. 


\textit{Architecture-based} methods in CL generate task-specific parameters by isolating subspaces or adding sub-networks \cite{redunet, acl, wsn, h^2, meat, ensemble, prog_net, path_net, rps_net}. However, these approaches need task identity during training and inference, making them unsuitable for task-free settings; also, they typically involve significant additional parameters \cite{den, learn-prune-share, mixed}.  In contrast, \cite{reviewer2-1} introduces virtual gradient updates from a virtual model, enabling ‘any-time inference’ for OCL. 

\textit{Regularization-based} methods selectively regularize the update of network parameters depending on their importance to the old tasks \cite{end-to-end, muc, large_scale-IL}. The importance of parameters can be determined using an approximation based on Fisher Information Matrix (FIM), as in EWC \cite{ewc}, Synaptic Intelligence \cite{SI} and MAS \cite{mas}. However, because EWC calculates the FIM at task transitions, it is not feasible in task-free OCL. On the other hand, EWC++ \cite{ewc++} combines the regularization approaches of EWC \cite{ewc} and Synaptic Intelligence \cite{SI} and makes it suitable for online settings. 


\textit{Rehearsal-based} methods address catastrophic forgetting by combining old training examples from a memory buffer with current data \cite{generative_replay, der, co2l, foster, gem, icarl, pcr, dsdm}. In principle, these methods are suitable for our task-free OCL setting, using strategies to retrieve and update the buffer \cite{mir, asv, gmed, dvc, aser, gss, cbrs, prs}. For instance, REMIND \cite{REMIND} enables efficient replay in OCL using compressed representations, while \cite{reviewer2-2} integrates rehearsal with regularization techniques. However, their effectiveness decreases with smaller buffers \cite{replay-buffer-size} and they pose challenges in data-sensitive environments \cite{data-privacy}


\textit{Prompt-based} methods construct a pool of task-specific prompts, select and attach them to the pre-trained model \cite{s-prompt, progressive-prompt, visual-prompt}. Most of these methods assume explicit task boundaries and require information on these task boundaries for training \cite{dualprompt, coda-prompt}; this is not feasible in task-free OCL. However, L2P \cite{l2p} is suitable for task-free OCL as it employs an instance-wise prompt query. Similarly, MVP \cite{mvp} is also suitable because it utilizes an instance-wise logit masking. In the class-incremental experiments within the original L2P codebase \cite{l2p}, a training trick is employed to mask out the classes not relevant to the current task. This trick contradicts the task-free OCL setting of having "no task identity information during training". Thus, to ensure a fair comparison, we evaluate our Online-LoRA against L2P \cite{l2p} and MVP \cite{mvp} under the Stochastic Incremental Blurry task boundary (Si-blurry) scenario, a new scenario introduced in the MVP paper \cite{mvp}.


\begin{figure*}[tb]
\centering
\includegraphics[width=0.95\textwidth]{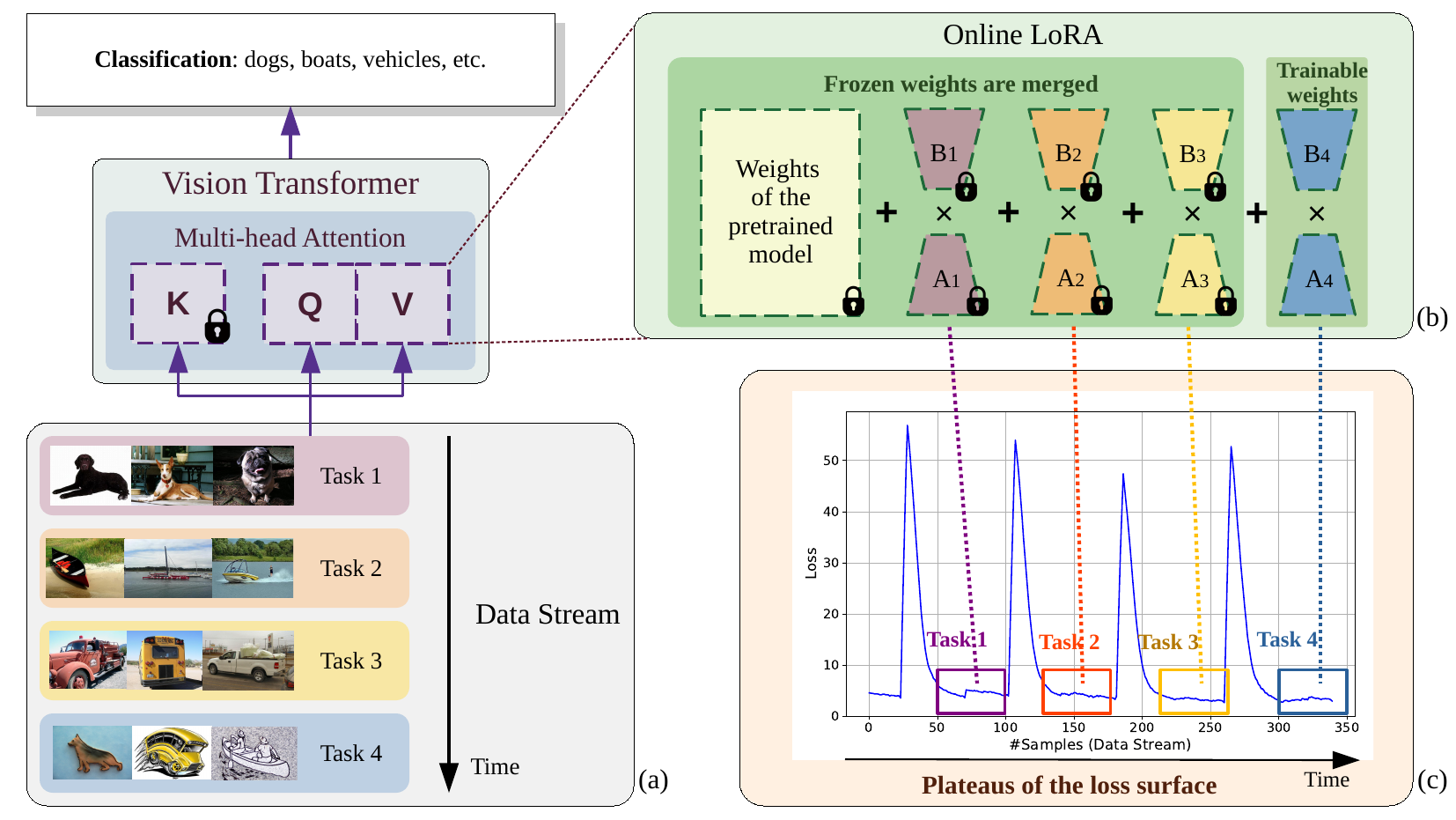}
\caption{The overview of Online-LoRA. As the data is continuously streamed (a), a new pair of trainable LoRA parameters ($A_4, B_4$) is added (b) every time the loss surface encounters a plateau (c). Subsequently, the previous LoRA parameters ($A_1, B_1; A_2, B_2; A_3, B_3$) are frozen (the lock sign in (b)) and merged to the weights of the pre-trained ViT model. }
\vspace{-10pt}
\label{fig:method}
\end{figure*}


\subsection{Parameter efficient fine-tuning}

Parameter Efficient Fine-Tuning (PEFT) is an effective approach for transfer learning \cite{peft-transfer}. Instead of fine-tuning an entire pre-trained model, PEFT fine-tunes specific sub-modules within the network by adding a small amount of additional parameters. PEFT reduces computation, but achieves similar performance to full fine-tuning. PEFT has been successfully applied to vision transformer models \cite{adaformer, ssf}, and one notable example is LoRA \cite{hu2021lora}. 

In LoRA, for a pre-trained weight matrix $W_{init}\in R^{d\times k}$, the update $\Delta W$ in $W\leftarrow W_{init}+\Delta W$ is reformulated as a low-rank decomposition: $\Delta W=BA$, where $A \in R^{r\times k}$ and $B \in R^{d\times r}$, and the rank $r \ll min(d, k)$. $W_{init}$ remains fixed during training and does not receive gradient updates, while A and B contain trainable parameters. 

The application of PEFT in transformer-based models has gained popularity in CL research \cite{color, construct-vl, task-arithmetic}. For example, SSIAT \cite{ssiat} incrementally tunes adapters \cite{adapter} in pre-trained ViT. Additionally, C-LoRA \cite{c-lora} and InfLoRA \cite{inf-lora} use separate LoRA sub-modules for each new task, and employ regularization to minimize interference between new and old tasks. However, these methods depend on the explicit knowledge of task boundaries, hence they are task-based offline CL approaches. To the best of our knowledge, our Online-LoRA is the first to extend LoRA to the task-free OCL scenario for transformer-based vision models.


\section{Online-LoRA}
\label{olora}


\subsection{Problem formulation}
\label{problem-formulation}
We define a data stream of unknown distributions $D = \{D_1 , . . . , D_N \}$ over $X \times Y$, where $X$ and $Y$ are input and output space respectively \cite{ocl-survey}. At each time step $s$, the system receives a batch of non i.i.d samples ${x_k^t , y_k^t}$ from the current distribution $D_t$ of task $t$; the system sees this batch only once. Moreover, at any moment $s$, the distribution $D_t$ can itself experience sudden or gradual changes from $D_t$ to $D_{t+1}$. The system is unaware of when and how these distribution changes happen. 

For simplicity, we assume that $D_t$ is the data distribution of task $t$, and any shift from $D_t$ to $D_{t+1}$ is sudden. Of note, this remains a task-free setting, since gradual transitions from $D_t$ to $D_{t+1}$ can still be modeled by adding intermediate tasks and making these distributions increasingly similar, thus effectively blurring the explicit boundaries between tasks. Our Online-LoRA does not assume any task boundaries at any time. 

\subsection{Loss-guided model adaptation} 
\label{main-plateau}

In existing LoRA-based CL methods \cite{c-lora, color}, new LoRA parameters are added for each new task $t'$, resulting in a set of LoRA parameters denoted as \{$A_{t'}, B_{t'}$\}, where $A_{t'} \in R^{d\times r}$ , $B_{t'} \in R^{r\times k}$ , $d$ and $k$ are the input and output dimensions of the attention layer, and rank $r \ll min(d, k)$. When learning task $t$, if the initial ViT weights are denoted as $W_{init}$, then for an input sample $X$, the model output $Y$ becomes: 
\begin{equation}
    Y = (W_{init} + \sum_{t'=0}^t B_{t'}A_{t'})X
\end{equation}

This \textit{incremental model} effectively mitigates the catastrophic forgetting by minimizing the interference between old and new tasks (see Figure~\ref{fig:method}). As shown, LoRA is applied only to the query and value projection matrices in all the attention layers. Since data from previous tasks is not available when training on future tasks, the LoRA parameters of old tasks are frozen and merged with the pre-trained weights to reduce the memory overhead. However, the existing LoRA-based methods rely on the knowledge of task boundaries during training, as a new pair of LoRA parameters is initialized at the beginning of each new task. In task-free OCL, data flows continuously without clear task boundaries, and there is no information about the start or the end of a task. There brings the need for a mechanism to determine when to initialize the new LoRA parameters.

To this end, we consider the idea of \textit{loss surface} \cite{mas_ocl}. As learning progresses, a decreasing loss indicates effective learning from current samples. Conversely, an increasing loss suggests a shift in data distribution, hindering effective learning. We assume that the model converges before the distribution shifts. Then between these phases, \textit{plateaus of the loss surface} occurs, signaling that the model has reached a stable state by fitting well to the current data distribution (see Appendix~\ref{sec:supp-loss_surface} for more details). At these plateaus, it is best to consolidate the learned knowledge by freezing the current LoRA weights and initializing a pair of new, trainable LoRA parameters. To prevent the accumulation of additional LoRA parameters, the frozen LoRA weights are merged into the pre-trained attention weights.

\subsection{Online parameter importance estimation} 
\label{regularization}

Many studies have demonstrated the efficacy of weight regularization in reducing catastrophic forgetting \cite{mas, ewc, ewc++}; this technique relies on estimating the importance of each parameter. However, in an online scenario where data distribution shifts constantly, parameter importance also varies continually. Therefore, a static estimation of parameter importance is not applicable. Furthermore, Online-LoRA utilizes pre-trained Vision Transformer (ViT) models, which have a \textit{substantial number of parameters}. Techniques such as calculating the Fisher information matrix at each task-switch for parameter importance estimation are computationally inefficient in this context \cite{ewc}. 

However, it is still useful to consider the model training from a Bayesian perspective as EWC does \cite{ewc}. In Bayesian machine learning, model parameters are treated as random variables, and the prior knowledge about these parameters is updated via Bayes' rule. More precisely, given data $D$:

\vspace{-15pt}
\begin{equation} 
\label{bayes}
    \log p(\theta|D) = \log p(D|\theta) + \log p(\theta) - \log p(D)
\end{equation}

Assume $D$ is split into two independent parts: current sample $x$ and data observed at the last time step $D_{prev}$. We can rewrite the posterior probability of the parameters and the equation (\ref{bayes}) becomes: 

\vspace{-10pt}
\begin{equation}
    \log p(\theta|D) = \log p(x|\theta) + \log p(\theta|D_{prev}) - \log p(x)
\end{equation}

Since calculating the posterior probability is usually intractable, following the work on the Laplace approximation \cite{laplace-approx}, we approximate this posterior as a Gaussian centered at the maximum a-posteriori (MAP) solution $\theta_{\text{MAP}}$ with covariance given by the inverse of the Hessian. In our work, the empirical Fisher information matrix is used to approximate the covariance of the approximated posterior. More specifically, the LoRA adapter $\sum_{t'} B_{t'}A_{t'}X$ is treated as two separate linear layers with weights $\sum_{t'} A_{t'}\in R^{d\times r}$ and $\sum_{t'} B_{t'}\in R^{r\times k}$, respectively, rather than as a single linear layer with a low rank weight matrix \cite{laplace-lora}; this division enhances memory efficiency. In EWC \cite{ewc}, the size of the importance weight matrix equals to the number of parameters squared. For instance, to employ EWC in ViT-B/16, the model needs to store and update a 86.6M$\times$86.6M matrix, representing a significant memory an computational overhead.  By handling the LoRA adapter as two distinct layers, our Online-LoRA approach employs two smaller importance weight matrices, $\Omega^{A,l} \in R^{d\times r}$ and $\Omega^{B,l} \in R^{r\times k}$, for each attention layer. The combined size of these matrices is proportional to the total number of LoRA parameters, calculated as follows: \#attention heads $\times$ 2 (for Q and V projection matrices) $\times$ input size $\times$ rank $\times$ 2. For a ViT-B/16 model with a LoRA rank of 4, this equates to a total of: 12 heads $\times$ 2 $\times$ 768 input size $\times$ 4 rank $\times$ 2 = 147,456. This additional memory footprint is negligible ($\sim$0.17\% of the total parameters of the ViT-B/16 model), which enables the \textit{online} updates of the importance weights. 

In offline CL, the parameter importance is computed based on the entire training dataset of the current task. This is not applicable to online CL because each training sample can only be seen once. We employ a small \textit{hard buffer} containing samples with the highest loss (computed with the current model), selected from both the current sample and the existing buffer. The hard buffer is continually updated to replace any samples whose loss decreases significantly as the model trains, ensuring that it contains genuinely challenging examples. Due to concerns about memory constraints and privacy, the hard buffer is minimal (holds only 4 samples), yet vital to parameter importance estimation.

Therefore, we propose a memory and computationally efficient estimation of the parameter importance, focusing on the sensitivity of loss relative to LoRA parameters. The \textit{importance weight matrices} $\Omega^{A,l} \in R^{d\times r} $ and $\Omega^{B,l} \in R^{r\times k}$ match the dimensions of LoRA parameters:
\vspace{-10pt}

\begin{align}
    \Omega^{A,l}_{ij} &= \frac{1}{N} \sum_{k=1}^{N}  \nabla_{W^{A,l}_{ij}} \log p(x_k | \theta) \circ\nabla_{W^{A,l}_{ij}} \log p(x_k | \theta)  \\
    \Omega^{B,l}_{ij} &= \frac{1}{N} \sum_{k=1}^{N}  \nabla_{W^{B,l}_{ij}} \log p(x_k | \theta) \circ \nabla_{W^{B,l}_{ij}} \log p(x_k | \theta)
\end{align}

with parameters $W^{A,l}$ and $W^{B,l}$ of the new and trainable LoRA modules added to the $l^{th}$ attention layer. $x_k$ are samples from the hard buffer described above. The parameters denoted by $\theta$ encompass the entire model, which includes the pre-trained ViT, the frozen LoRA parameters, and the new trainable LoRA parameters $W^{A,l}$ and $W^{B,l}$ for all attention layers $l$. Finally, $\circ$ is the Hadamard product (i.e. the element-wise product of two matrices). 

After updating the importance weights, the model continues the learning process while penalizing changes to parameters that have been identified as important so far. Because $\theta_{\text{MAP}} = \underset{\theta}{\operatorname{argmax}} \, \mathcal{L}(y, D; \theta)$, $\theta_{\text{MAP}}$ is given by model weights at the last loss surface plateau. As such, our final learning objective is: 
\vspace{-5pt}
\begin{equation}
\label{main-final-objective}
    \min_{W^A, W^B} \mathcal{L}(F(X; \theta), Y) + \mathcal{L}(F(X_B; \theta), Y_B) + L_\text{LoRA}(W^A, W^B)
\end{equation}
\vspace{-20pt}

\begin{align}
\label{main-lora-loss}
    L_\text{LoRA}(W^A, W^B) &= \frac{\lambda}{2} \sum_{l \in |Attn|} ((\Omega^{A,l}\circ (W^{A,l})\circ (W^{A,l})) \notag \\
    &\quad + (\Omega^{B,l}\circ (W^{B,l})\circ (W^{B,l})))
\end{align}

where $Attn$ is the set of attention layers in the model, $\mathcal{L}(F(X; \theta), Y)$ is the loss of current samples, $\mathcal{L}(F(X_B; \theta), Y_B)$ is the loss of hard buffer samples. $\circ$ is the element-wise product of two matrices. 

\section{Experiments}
\label{main-exps}

\begin{table*}[t!]
\centering
\begin{adjustbox}{max width=\textwidth}
\setlength{\tabcolsep}{6pt}
\begin{tabular}{lccccccccc}
\toprule
 &  & \multicolumn{2}{c}{Split-CIFAR-100} & \multicolumn{2}{c}{Split-ImageNet-R} & \multicolumn{2}{c}{Split-ImageNet-S} & \multicolumn{2}{c}{Split-CUB-200} \\
\cmidrule(lr){3-4} \cmidrule(lr){5-6} \cmidrule(lr){7-8} \cmidrule(lr){9-10}
 & & $A_{\text{Final}}$ ($\uparrow$) & \textit{Forgetting} ($\downarrow$) & $A_{\text{Final}}$ ($\uparrow$) & \textit{Forgetting} ($\downarrow$) & $A_{\text{Final}}$ ($\uparrow$) & \textit{Forgetting} ($\downarrow$) & $A_{\text{Final}}$ ($\uparrow$) & \textit{Forgetting} ($\downarrow$) \\
\midrule
\multirow{12}{*}{ViT-B/16} & AGEM \cite{a-gem} & 12.67\scriptsize$\pm$1.87 & 82.51\scriptsize$\pm$2.27 & 5.60\scriptsize$\pm$2.74 & 53.97\scriptsize$\pm$1.97 & 0.16\scriptsize$\pm$0.04 & 9.42\scriptsize$\pm$0.17 & 10.84\scriptsize$\pm$1.57 & 47.79\scriptsize$\pm$0.04 \\
 & ER \cite{er} & 44.85\scriptsize$\pm$1.83 & 44.67\scriptsize$\pm$4.29 & 40.99\scriptsize$\pm$3.96 & 32.38\scriptsize$\pm$0.89 & 30.21\scriptsize$\pm$0.70 & 37.14\scriptsize$\pm$1.83 & 31.66\scriptsize$\pm$0.83 & 14.23\scriptsize$\pm$0.07 \\
 & EWC++ \cite{ewc++} & 10.61\scriptsize$\pm$0.74 & 84.10\scriptsize$\pm$1.11 & 3.86\scriptsize$\pm$2.02 & 56.95\scriptsize$\pm$1.46 & 0.32\scriptsize$\pm$0.28 & 22.46\scriptsize$\pm$4.69 & 26.14\scriptsize$\pm$3.46 & 47.69\scriptsize$\pm$0.07 \\
 & MIR \cite{mir} & 48.36\scriptsize$\pm$3.11 & 43.41\scriptsize$\pm$1.02 & 41.51\scriptsize$\pm$2.99 & 31.32\scriptsize$\pm$5.17 & 30.33\scriptsize$\pm$3.81 & 35.92\scriptsize$\pm$1.75 & 31.64\scriptsize$\pm$2.97 & 23.43\scriptsize$\pm$0.05 \\
 & GDumb \cite{gdumb} & 41.00\scriptsize$\pm$19.97 & - & 8.87\scriptsize$\pm$1.36 & - & 1.65\scriptsize$\pm$0.22 & - & 9.09\scriptsize$\pm$1.03 & - \\
 & PCR \cite{pcr} & 48.48\scriptsize$\pm$0.15 & 46.23\scriptsize$\pm$1.29 & 46.11\scriptsize$\pm$3.03 & 25.50\scriptsize$\pm$0.41 & 38.75\scriptsize$\pm$0.22 & 35.01\scriptsize$\pm$2.12 & 41.11\scriptsize$\pm$1.43 & 29.64\scriptsize$\pm$1.20 \\
 & DER++ \cite{der} & 36.64\scriptsize$\pm$6.11 & 56.94\scriptsize$\pm$7.55 & 30.90\scriptsize$\pm$8.04 & 24.26\scriptsize$\pm$4.14 & 6.47\scriptsize$\pm$0.06 & 15.34\scriptsize$\pm$0.15 & 26.61\scriptsize$\pm$1.27 & 32.16\scriptsize$\pm$0.55 \\
 & LODE (DER++) \cite{lode} & 44.29\scriptsize$\pm$1.48 & 45.54\scriptsize$\pm$3.32 & 42.20\scriptsize$\pm$6.46 & 31.83\scriptsize$\pm$1.05 & 9.97\scriptsize$\pm$2.29 & \textbf{8.48\scriptsize$\pm$1.24} & 39.20\scriptsize$\pm$4.25 & 41.64\scriptsize$\pm$3.59 \\
 & EMA (DER++) \cite{online-ema} & 42.28\scriptsize$\pm$4.36 & 55.59\scriptsize$\pm$1.48 & 41.75\scriptsize$\pm$1.98 & 32.65\scriptsize$\pm$1.55 & 16.88\scriptsize$\pm$2.23 & 36.28\scriptsize$\pm$1.09 & 35.26\scriptsize$\pm$3.31 & 25.55\scriptsize$\pm$3.35 \\
 & EMA (RAR) \cite{online-ema} & 47.10\scriptsize$\pm$0.82 & 50.01\scriptsize$\pm$0.35 & 30.04\scriptsize$\pm$0.33 & 39.36\scriptsize$\pm$0.04 & 14.06\scriptsize$\pm$0.37 & 36.28\scriptsize$\pm$1.09 & 33.34\scriptsize$\pm$1.11 & 28.68\scriptsize$\pm$0.56 \\
 & \textbf{Ours} & \textbf{49.40\scriptsize$\pm$1.36} & \textbf{41.74\scriptsize$\pm$2.58} & \textbf{48.18\scriptsize$\pm$0.63} & \textbf{23.85\scriptsize$\pm$1.48} & \textbf{47.06\scriptsize$\pm$0.24} & 28.09\scriptsize$\pm$3.25 & \textbf{41.46\scriptsize$\pm$0.31} & \textbf{13.64\scriptsize$\pm$0.68} \\
 & \textit{UB} & 89.50\scriptsize$\pm$0.04 & - & 76.78\scriptsize$\pm$0.44 & - & 63.82\scriptsize$\pm$0.02 & - & 82.81\scriptsize$\pm$1.07 & - \\
\midrule
\multirow{12}{*}{ViT-S/16} & AGEM \cite{a-gem} & 7.43\scriptsize$\pm$2.15 & 82.45\scriptsize$\pm$5.46 & 2.35\scriptsize$\pm$0.87 & 48.01\scriptsize$\pm$0.05 & 2.75\scriptsize$\pm$2.86 & \textbf{18.81\scriptsize$\pm$0.44} & 1.40\scriptsize$\pm$0.17 & 27.06\scriptsize$\pm$1.39 \\
 & ER \cite{er} & 31.91\scriptsize$\pm$2.06 & 52.21\scriptsize$\pm$6.41 & 32.73\scriptsize$\pm$0.20 & 45.37\scriptsize$\pm$1.72 & 19.53\scriptsize$\pm$1.44 & 45.10\scriptsize$\pm$0.48 & 21.81\scriptsize$\pm$3.02 & 24.52\scriptsize$\pm$2.98 \\
 & EWC++ \cite{ewc++} & 6.80\scriptsize$\pm$2.13 & 81.59\scriptsize$\pm$7.43 & 1.32\scriptsize$\pm$0.83 & 53.54\scriptsize$\pm$0.19 & 4.08\scriptsize$\pm$3.24 & 21.28\scriptsize$\pm$0.46 & 2.07\scriptsize$\pm$0.54 & 28.44\scriptsize$\pm$0.83 \\
 & MIR \cite{mir} & 29.08\scriptsize$\pm$1.14 & 39.42\scriptsize$\pm$1.60 & \textbf{34.73\scriptsize$\pm$0.29} & 48.66\scriptsize$\pm$0.69 & 13.96\scriptsize$\pm$1.95 & 42.61\scriptsize$\pm$0.08 & 22.95\scriptsize$\pm$1.12 & 32.54\scriptsize$\pm$0.88 \\
 & GDumb \cite{gdumb} & 10.87\scriptsize$\pm$4.94 & - & 5.33\scriptsize$\pm$1.09 & - & 2.09\scriptsize$\pm$0.32 & - & 3.28\scriptsize$\pm$0.99 & - \\
 & PCR \cite{pcr} & 32.89\scriptsize$\pm$1.47 & 39.90\scriptsize$\pm$2.51 & 21.96\scriptsize$\pm$0.27 & 45.12\scriptsize$\pm$0.08 & 14.37\scriptsize$\pm$0.95 & 43.96\scriptsize$\pm$0.48 & 22.28\scriptsize$\pm$2.73 & 29.87\scriptsize$\pm$0.04 \\
 & DER++ \cite{der} & 17.67\scriptsize$\pm$4.04 & 51.65\scriptsize$\pm$3.67 & 22.17\scriptsize$\pm$4.27 & 54.79\scriptsize$\pm$0.89 & 18.15\scriptsize$\pm$0.66 & 46.22\scriptsize$\pm$0.95 & 29.53\scriptsize$\pm$2.37 & 21.49\scriptsize$\pm$1.16 \\
 & LODE (DER++) \cite{lode} & 28.65\scriptsize$\pm$3.06 & 40.42\scriptsize$\pm$1.58 & 31.65\scriptsize$\pm$0.72 & 43.72\scriptsize$\pm$0.09 & 17.59\scriptsize$\pm$0.84 & 47.85\scriptsize$\pm$0.23 & 26.81\scriptsize$\pm$0.89 & 21.86\scriptsize$\pm$2.30 \\
 & EMA (DER++) \cite{online-ema} & 12.76\scriptsize$\pm$0.65 & 41.17\scriptsize$\pm$1.75 & 20.89\scriptsize$\pm$3.05 & 48.03\scriptsize$\pm$1.79 & 12.93\scriptsize$\pm$0.13 & 22.59\scriptsize$\pm$0.16 & 35.79\scriptsize$\pm$5.27 & 24.85\scriptsize$\pm$4.20 \\
 & EMA (RAR) \cite{online-ema} & 19.21\scriptsize$\pm$2.16 & 41.99\scriptsize$\pm$1.73 & 16.11\scriptsize$\pm$0.35 & 50.58\scriptsize$\pm$0.83 & 14.50\scriptsize$\pm$2.71 & 23.79\scriptsize$\pm$2.91 & 34.53\scriptsize$\pm$1.04 & 30.19\scriptsize$\pm$0.36 \\
 & \textbf{Ours} & \textbf{32.16\scriptsize$\pm$0.24} & \textbf{38.64\scriptsize$\pm$0.65} & 33.21\scriptsize$\pm$0.50 & \textbf{42.76\scriptsize$\pm$0.18} & \textbf{22.45\scriptsize$\pm$0.43} & 44.56\scriptsize$\pm$0.24 & \textbf{37.41\scriptsize$\pm$0.16} & \textbf{20.78\scriptsize$\pm$2.54} \\
 & \textit{UB} & 86.55\scriptsize$\pm$0.01 & - & 69.94\scriptsize$\pm$0.34 & - & 59.28\scriptsize$\pm$0.11 & - & 73.91\scriptsize$\pm$1.16 & - \\
\bottomrule
\vspace{-20pt}
\end{tabular}
\end{adjustbox}
\caption{Results of disjoint class-incremental learning. `$\uparrow$' means higher is better and `$\downarrow$' means lower is better. Regularization-based methods (EWC++, AGEM, and LODE) yield low accuracy and low forgetting on Split ImageNet-S. This is because their overly rigid constraints on model updates hinder effective learning. The best results are noted by \textbf{bold}. \textit{UB} is the upper-bound performance. }
\label{tab:disjoint cil}
\end{table*}

\subsection{Evaluation benchmarks}
\label{benchmark}

We evaluate our approach under three different scenarios: disjoint class-incremental, Si-Blurry class-incremental, and domain-incremental. 

\textbf{Disjoint class-incremental} setting is when the datasets are split into disjoint tasks, each consisting of a unique set of classes. We conduct experiments with three datasets under this setting: Split-CIFAR-100 splits the CIFAR-100 dataset \cite{cifar100} into 10 tasks with 10 classes per task. Split-ImageNet-R splits the ImageNet-R dataset \cite{imagenet-r} into 10 tasks with 20 classes per task. Split-ImageNet-S splits the ImageNet-Sketch dataset \cite{imagenet-s} randomly into 10 tasks with 100 classes per task or into 20 tasks with 50 classes per task. Split-CUB-200 splits the CUB-200-2011 dataset \cite{cub-200} into 5 tasks with 40 classes per task. 

\textbf{Stochastic incremental-Blurry (Si-Blurry) \cite{si-blurry} class-incremental} setting is when the class distributions change in a stochastic manner, with classes overlapping across tasks and the task boundaries being dynamic and not clearly defined. We randomly select 50\% of the entire classes to be "disjoint classes" (newly encountered classes that never appeared before), and 10\% to be "blurry classes" (classes that do not belong to a fixed task and may appear in multiple learning tasks over time). 

\textbf{Domain-incremental} setting is when the input distribution shifts over time, but the classes remain consistent. We use the CORe50 dataset \cite{core50} for this setting; it has 11 distinct domains (8 for training, 3 for testing). The samples from the training domains arrive sequentially. 

\begin{table*}[t!]
\centering
\begin{adjustbox}{max width=\textwidth}
\begin{tabular}{cccccccc}
\toprule
&  & \multicolumn{2}{c}{CIFAR-100 \cite{cifar100}} &
  \multicolumn{2}{c}{ImageNet-R \cite{imagenet-r}} &
  \multicolumn{2}{c}{ImageNet-S \cite{imagenet-s}} \\
&  & {$A_{\text{AUC}}$ ($\uparrow$)} & {$A_{\text{Final}}$ ($\uparrow$)} & {$A_{\text{AUC}}$ ($\uparrow$)} & {$A_{\text{Final}}$ ($\uparrow$)} & {$A_{\text{AUC}}$ ($\uparrow$)} & {$A_{\text{Final}}$ ($\uparrow$)} \\
\cmidrule(lr){3-4}\cmidrule(lr){5-6}\cmidrule(lr){7-8}
\multirow{4}{*}{ViT-B/16} & L2P  & 43.01\scriptsize$\pm$9.37 & 39.86\scriptsize$\pm$2.28 & 22.71\scriptsize$\pm$1.86 & 27.08\scriptsize$\pm$2.49 & 10.02\scriptsize$\pm$0.42 & 13.58\scriptsize$\pm$4.04 \\
 & MVP & 47.52\scriptsize$\pm$9.74 & 44.49\scriptsize$\pm$0.93 & 27.79\scriptsize$\pm$0.62 & 31.64\scriptsize$\pm$1.77 & 10.68\scriptsize$\pm$0.45 & 13.99\scriptsize$\pm$1.73 \\
 & \textbf{Ours} &  \textbf{60.12\scriptsize$\pm$5.79} & \textbf{61.70\scriptsize$\pm$6.29} & \textbf{45.05\scriptsize$\pm$1.59} & \textbf{48.00\scriptsize$\pm$6.01} & \textbf{30.81\scriptsize$\pm$2.09} & \textbf{30.22\scriptsize$\pm$4.36}\\
 & \textit{UB} & \multicolumn{2}{c}{89.50\scriptsize$\pm$0.04} & \multicolumn{2}{c}{76.78\scriptsize$\pm$0.44} & \multicolumn{2}{c}{63.82\scriptsize$\pm$0.02}\\
\midrule
\multirow{4}{*}{ViT-S/16} & L2P & 37.82\scriptsize$\pm$12.19 & 30.88\scriptsize$\pm$1.39 & 24.31\scriptsize$\pm$1.83 & 21.83\scriptsize$\pm$2.13 & 2.00\scriptsize$\pm$0.12 & 3.61\scriptsize$\pm$1.08 \\
& MVP & 40.31\scriptsize$\pm$9.52 & 35.55\scriptsize$\pm$2.11 & 27.04\scriptsize$\pm$1.09 & 26.67\scriptsize$\pm$3.70 & 2.27\scriptsize$\pm$0.14 & 3.72\scriptsize$\pm$0.77 \\
 & \textbf{Ours} &\textbf{52.84\scriptsize$\pm$7.97} & \textbf{58.72\scriptsize$\pm$1.44} & \textbf{39.47\scriptsize$\pm$1.93} & \textbf{36.61\scriptsize$\pm$4.63} & \textbf{15.35\scriptsize$\pm$0.92} & \textbf{20.18\scriptsize$\pm$1.84}\\
& \textit{UB} & \multicolumn{2}{c}{86.55\scriptsize$\pm$0.01} & \multicolumn{2}{c}{69.94\scriptsize$\pm$0.34} & \multicolumn{2}{c}{59.28\scriptsize$\pm$0.11}\\
\bottomrule
\vspace{-20pt}
\end{tabular}
\end{adjustbox}
\caption{Results of Si-blurry class-incremental learning. `$\uparrow$' means higher is better and `$\downarrow$' means lower is better. All datasets are split into 5 blurry tasks. To ensure a fair comparison with L2P \cite{l2p} and MVP \cite{mvp}, we exclude the loss from hard buffer samples in Online-LoRA. The best results are noted by \textbf{bold}.}
\vspace{-13pt}
\label{table:blur cil}
\end{table*}


\subsection{Experimental details}
\label{main-exp-detail}

\paragraph{Baselines} We compare Online-LoRA against SOTA task-free OCL methods. The Upper-bound (\textit{UB}) baseline refers to supervised fine-tuning on the entire dataset of i.i.d. data, representing the optimal performance. The SOTA methods selected for comparison include AGEM \cite{a-gem}, ER \cite{er}, EWC++ \cite{ewc++}, MIR \cite{mir}, GDumb \cite{gdumb}, DER++ \cite{der}, PCR \cite{pcr}, LODE (with DER++ \cite{der}) \cite{lode}, EMA (with DER++ \cite{der} and with RAR \cite{rar}) \cite{online-ema}, L2P \cite{l2p} and MVP \cite{mvp}. 

\vspace{-14pt}
\paragraph{Evaluation metrics} To evaluate the OCL performance, we choose three metrics, $A_{\text{AUC}}$, $A_{\text{Final}}$, and \textit{Forgetting}. The $A_{\text{AUC}}$ \cite{auc} evaluates the model accuracy throughout training, measuring the performance of anytime inference. The final accuracy $A_{\text{Final}}$ \cite{a-last, scr} measures the performance after the training is finished. \textit{Forgetting} \cite{l2p} measures the average difference between the final performance obtained for each task compared to the best performance on each task. Higher $A_{\text{AUC}}$ and $A_{\text{Final}}$ are better, while lower $Forgetting$ is better. See Appendix~\ref{sec:supp-metrics} for the detailed definitions. 

\vspace{-14pt}
\paragraph{Implementation details} We employ a ViT-B/16 (86.6M parameters) and a ViT-S/16 (48.6M parameters) \cite{vit} pre-trained on ImageNet as our backbone. For each setup, we evaluate all methods, including ours and other SOTA methods, using the same pre-trained models (see Appendix~\ref{sec: supp-baseline-backbone}).

We use the Adam optimizer \cite{adam-opt} to train our Online-LoRA, with a 0.0002 learning rate for ViT-B/16 and 0.0005 for ViT-S/16. We set the size of the minimal hard buffer to 4, regularization factor $\lambda$ to 2000 for all settings. See Appendix~\ref{sec:supp-exp_detail} for experimental details of Online-LoRA. For the other approaches, we refer to their original codebases for implementation and hyperparameter selection for a fair comparison (details in Appendix~\ref{sec:supp-baseline-setting}). The buffer size of the replay-based methods is 500 (results for other buffer sizes in Appendix~\ref{sec:supp-buffer-size}). Given our focus on online CL, the training epoch is set to 1 for all experiments.


\subsection{Main results}
\label{sec:main-results}

\textbf{Results on disjoint class-incremental setting}. Table~\ref{tab:disjoint cil} summarizes the results on the disjoint class-incremental benchmarks Split CIFAR-100, Split ImageNet-R, Split ImageNet-S, and Split CUB-200. Our Online-LoRA, outperforms all other compared methods consistently across the ViT-B/16 and ViT-S/16. On Split ImageNet-S, Online-LoRA exhibits standout performance, significantly outperforming all other methods, and notably reducing the gap to the upper bound. 

As shown in Table~\ref{tab:disjoint cil}, Online-LoRA maintains a consistent and strong performance across various dataset sizes. In comparison, GDumb \cite{gdumb} exhibits unstable performance on the smaller dataset, Split CIFAR-100, and performs poorly on larger datasets such as Split-ImageNet-R and Split ImageNet-S. The main issue with GDumb is its exclusive reliance on a replay buffer to retrain the model. With larger datasets, a small buffer size tends to cause class imbalance, as it cannot represent the dataset diversity adequately. Online-LoRA, on the other hand, does not face this issue because it utilizes a small but highly targeted 'hard buffer' consisting of samples that the current model finds most challenging, as indicated by their high loss values. This selective buffering approach is not only effective, as shown in Section~\ref{sec:main-ablation_study}, but it also sidesteps the drawbacks of a large memory buffer by not overly relying on it.

\begin{figure*}[tb]
\centering
\begin{subfigure}{.33\textwidth}
  \centering
  \includegraphics[width=\linewidth]{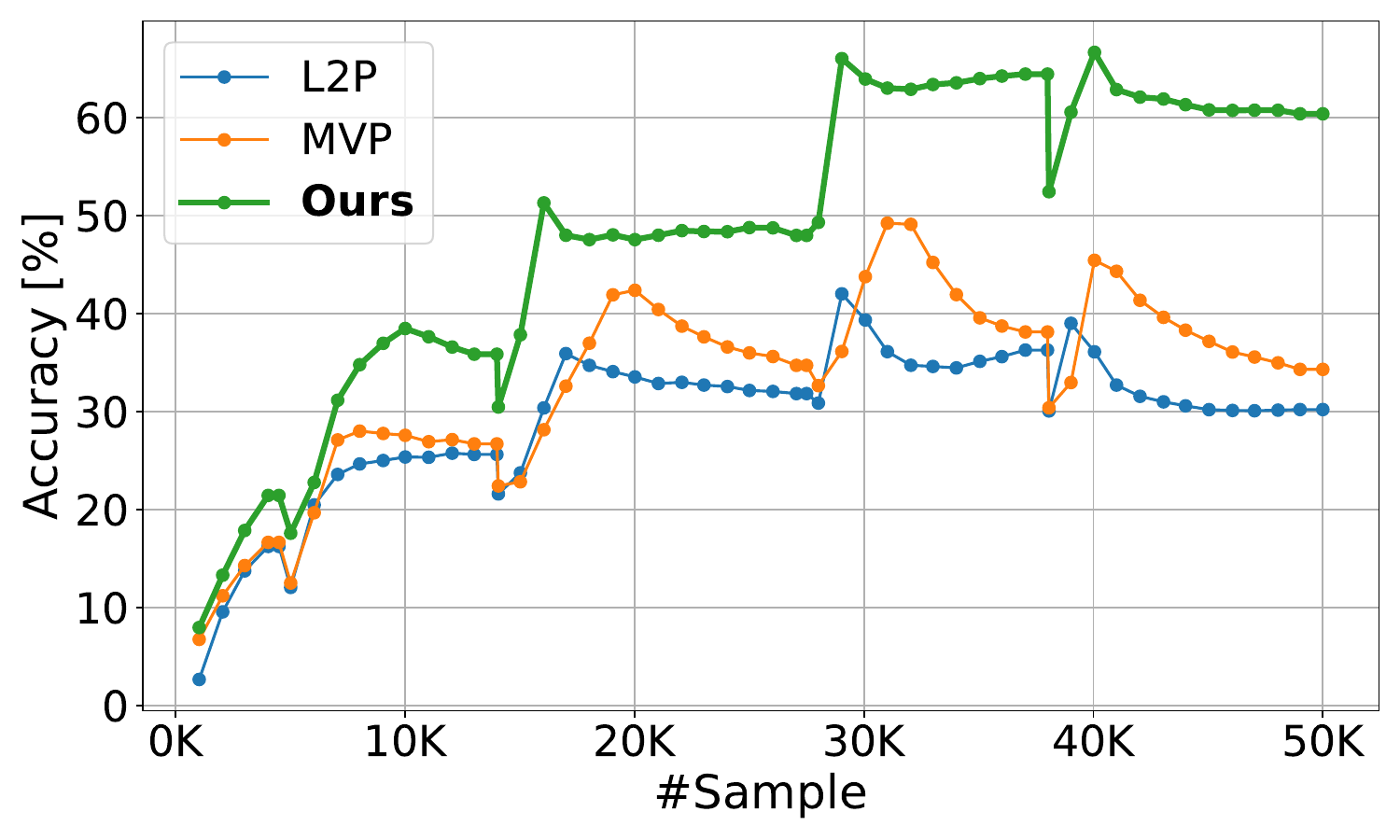}
  \caption{CIFAR-100 with ViT-B/16}
  \label{fig:cifar_base_blur}
\end{subfigure}%
\begin{subfigure}{.33\textwidth}
  \centering
  \includegraphics[width=\linewidth]{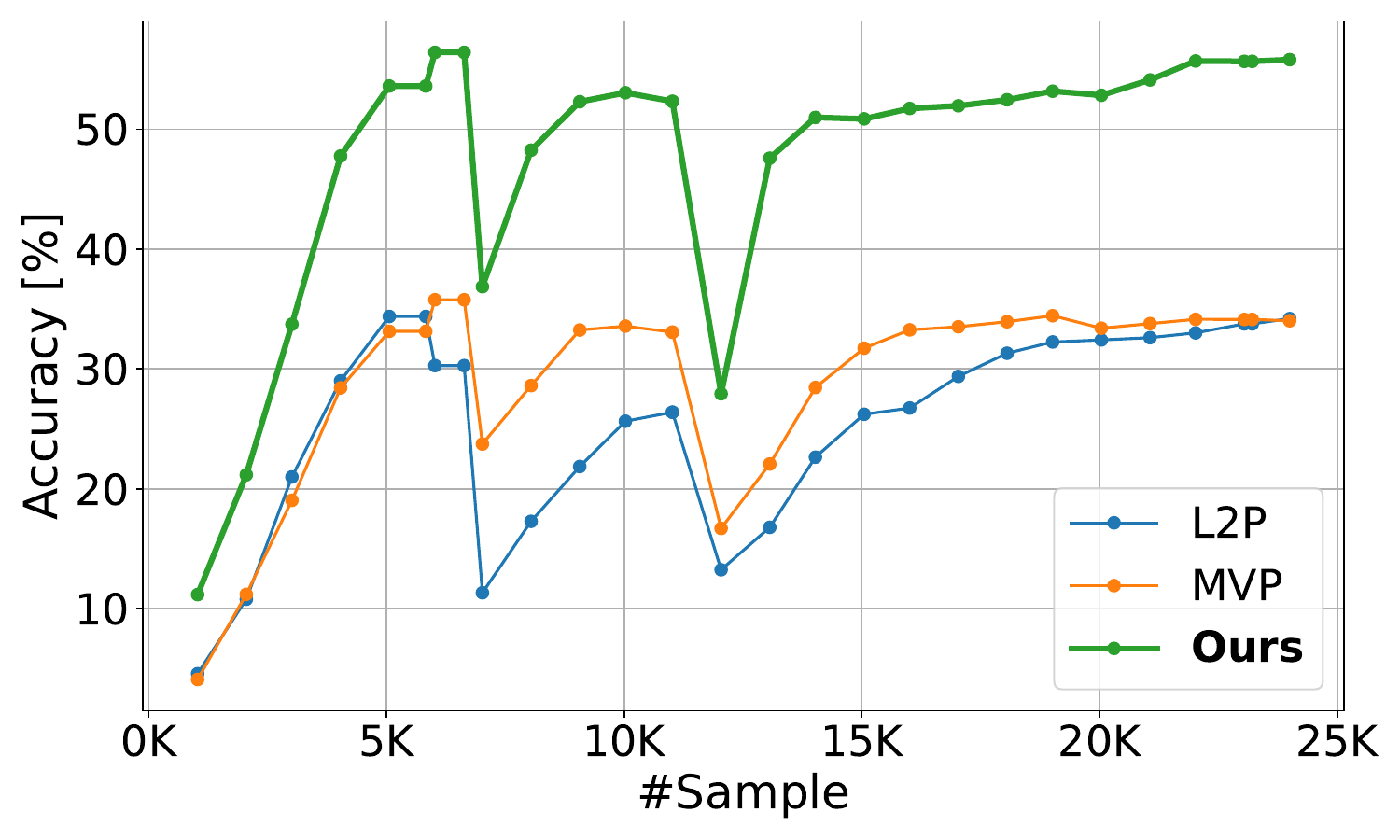}
  \caption{ImageNet-R with ViT-B/16}
  \label{fig:imnR_base_blur}
\end{subfigure}%
\begin{subfigure}{.33\textwidth}
  \centering
  \includegraphics[width=\linewidth]{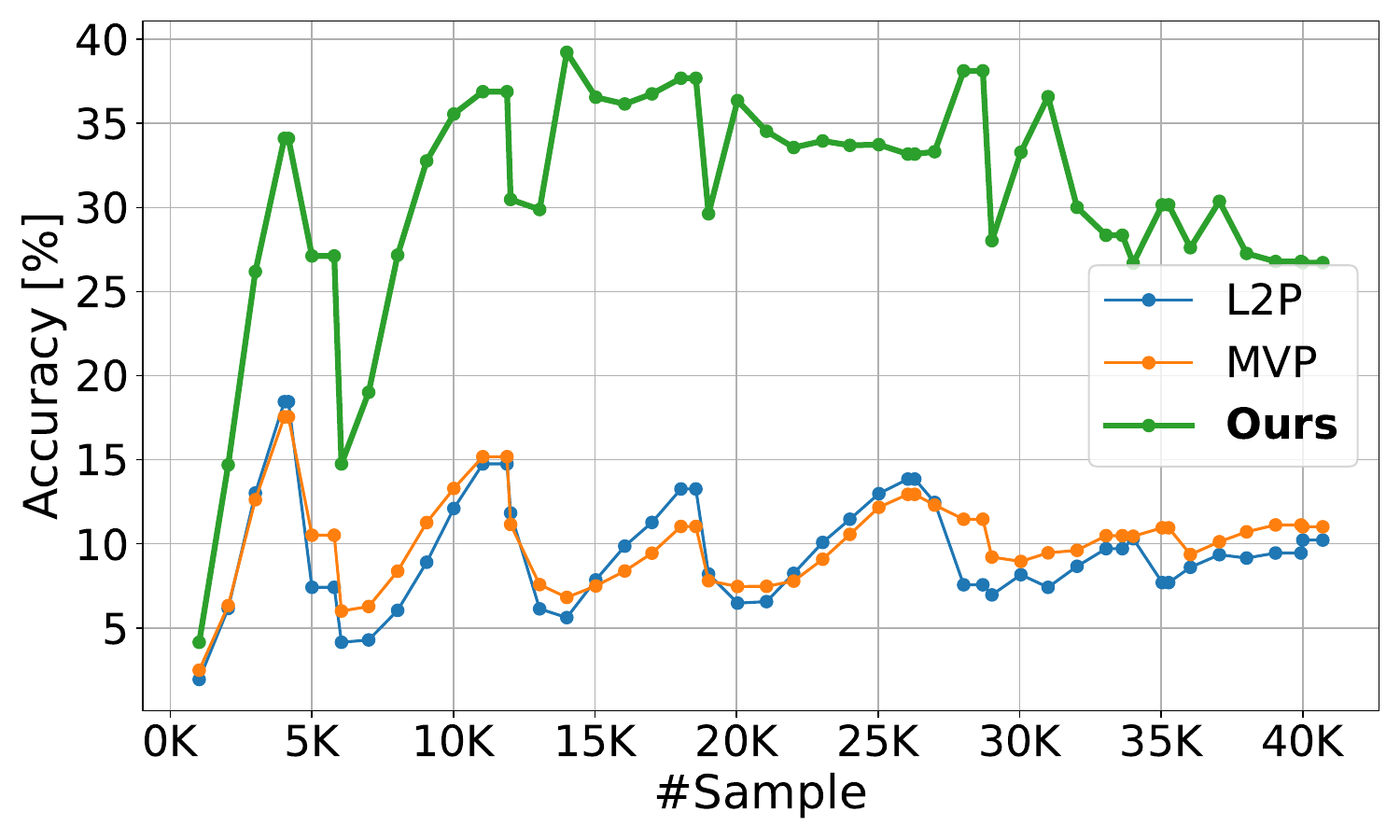}
  \caption{ImageNet-S with ViT-B/16}
  \label{fig:sketch_base_blur}
\end{subfigure}
\begin{subfigure}{.33\textwidth}
  \centering
  \includegraphics[width=\linewidth]{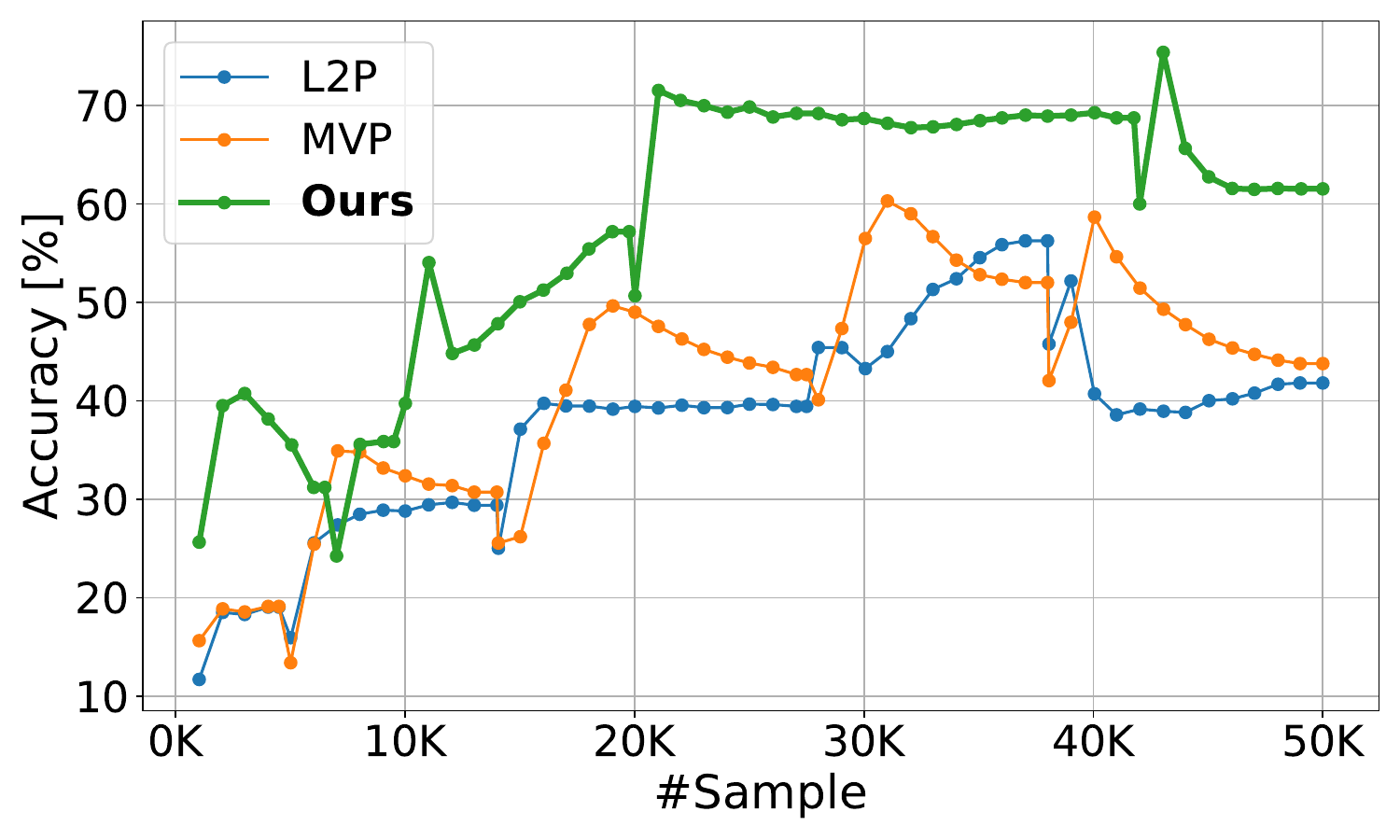}
  \caption{CIFAR-100 with ViT-S/16}
  \label{fig:cifar_small_blur}
\end{subfigure}%
\begin{subfigure}{.33\textwidth}
  \centering
  \includegraphics[width=\linewidth]{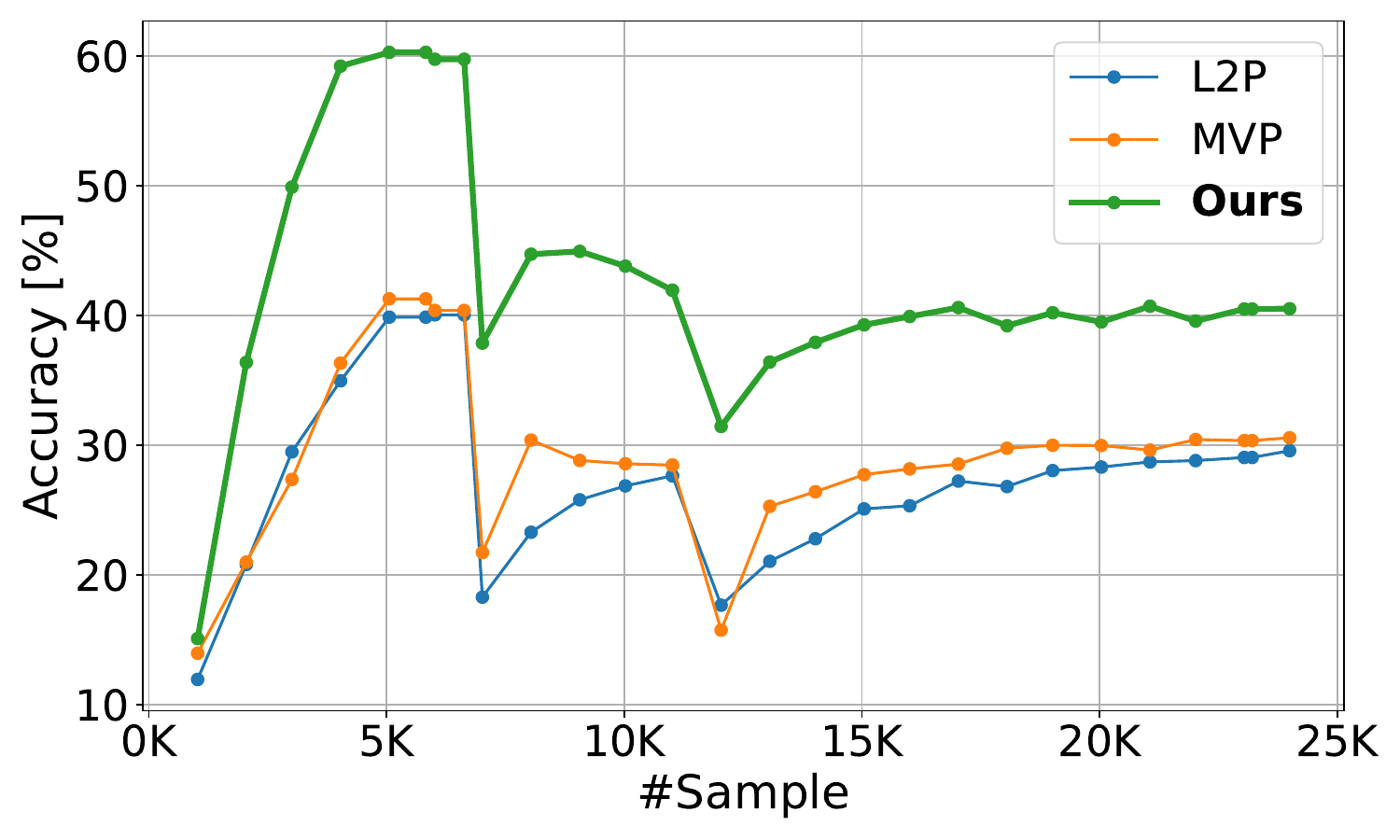}
  \caption{ImageNet-R with ViT-S/16}
  \label{fig:imnR_small_blur}
\end{subfigure}%
\begin{subfigure}{.33\textwidth}
  \centering
  \includegraphics[width=\linewidth]{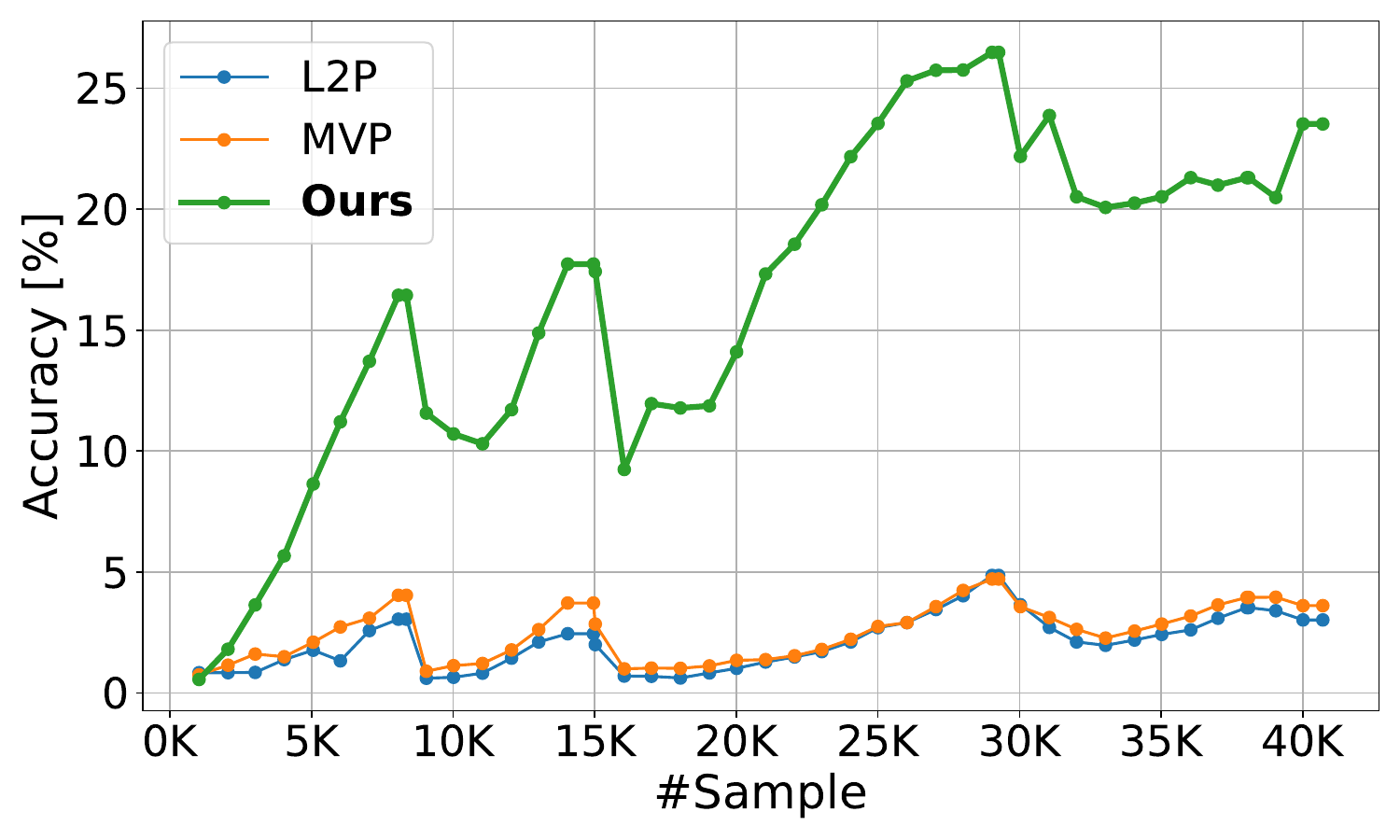}
  \caption{ImageNet-S with ViT-S/16}
  \label{fig:sketch_small_blur}
\end{subfigure}
\vspace{-8pt}
\caption{Average accuracy versus number of samples for Si-Blurry CIFAR-100, ImageNet-R, and ImageNet-S scenarios. As shown, the Online-LoRA consistently outperforms competing methods, maintaining high accuracy throughout.}
\vspace{-5pt}
\label{fig:acc_blur}
\end{figure*}

\textbf{Results on Si-blurry class-incremental setting}. Table~\ref{table:blur cil} summarizes the results on the Si-blurry class-incremental benchmarks with datasets CIFAR-100, ImageNet-R, and ImageNet-S. In the Si-blurry scenario, Online-LoRA consistently outperforms all the considered methods by significant margins across both metrics, $A_{\text{AUC}}$ and $A_{\text{Final}}$. The superior performance in anytime inference can be largely attributed to Online-LoRA strategic utilization of loss surface plateaus, which consolidates the knowledge precisely when needed. Online-LoRA is also more flexible than EWC \cite{ewc} which does so only at specific discrete moments; Online-LoRA also avoids the excessive frequency of updates that introduce noise as seen in EWC++ \cite{ewc++}. 

Figure~\ref{fig:acc_blur} displays the trend of accuracy as more samples are provided, highlighting the consistent performance of Online-LoRA across two different ViT architectures. Compared to other methods, Online-LoRA effectively learns new knowledge from incoming samples, which leads to an increase in accuracy.

\textbf{Results on domain-incremental setting}. Table~\ref{table:dil} summarizes the results on the domain-incremental setting. Our proposed method, Online-LoRA, not only significantly outperforms other SOTA methods, but also closes a substantial portion of the gap with the upper-bound (UB) performance.

To summarize, the Online-LoRA consistently achieves superior performance under various setups. These results indicate its robustness and adaptability, not only in different ViT setups, but also for dynamically evolving data. In addition to effectively mitigating forgetting, Online-LoRA shows good plasticity. 


\begin{table}[tb]
\centering
\begin{adjustbox}{max width=\linewidth}
\begin{tabular}{cccccc}
\toprule
 & \multicolumn{2}{c}{ViT-B/16} & \multicolumn{2}{c}{ViT-S/16} \\
\cmidrule(lr){2-3} \cmidrule(lr){4-5}
 & $A_{\text{Final}}$ ($\uparrow$) & \textit{Forgetting} ($\downarrow$) & $A_{\text{Final}}$ ($\uparrow$) & \textit{Forgetting} ($\downarrow$) \\
\midrule
AGEM \cite{a-gem} & 80.15\scriptsize$\pm$2.97 & 2.23\scriptsize$\pm$0.81 & 78.22\scriptsize$\pm$3.51 & 3.19\scriptsize$\pm$0.09 \\
ER \cite{er} & 85.85\scriptsize$\pm$1.35 & 0.72\scriptsize$\pm$0.03 & 78.99\scriptsize$\pm$3.85 & 5.04\scriptsize$\pm$0.10 \\
EWC++ \cite{ewc++} & 78.65\scriptsize$\pm$6.51 & 2.31\scriptsize$\pm$0.17 & 79.03\scriptsize$\pm$4.54 & 4.80\scriptsize$\pm$0.69 \\
MIR \cite{mir} & 74.35\scriptsize$\pm$4.07 & 11.01\scriptsize$\pm$1.05 & 86.49\scriptsize$\pm$0.81 & 2.53\scriptsize$\pm$0.84 \\
GDumb \cite{gdumb} & 77.20\scriptsize$\pm$3.49 & - & 75.64\scriptsize$\pm$2.92 & - \\
PCR \cite{pcr} & 87.16\scriptsize$\pm$0.73 & 0.78\scriptsize$\pm$0.03 & 75.20\scriptsize$\pm$1.48 & 0.61\scriptsize$\pm$0.02 \\
DER++ \cite{der} & 81.88\scriptsize$\pm$7.06 & 10.13\scriptsize$\pm$7.00 & 89.33\scriptsize$\pm$0.62 & 0.42\scriptsize$\pm$0.57 \\
LODE (DER++) \cite{lode} & 77.02\scriptsize$\pm$2.22 & 17.30\scriptsize$\pm$2.82 & 83.48\scriptsize$\pm$5.84 & 24.54\scriptsize$\pm$0.94 \\
L2P \cite{l2p} & 87.97\scriptsize$\pm$0.37 & \textbf{0.00\scriptsize$\pm$0.00} & 86.47\scriptsize$\pm$0.23 & \textbf{0.00\scriptsize$\pm$0.00} \\
MVP \cite{mvp} & 84.82\scriptsize$\pm$0.54 & \textbf{0.00\scriptsize$\pm$0.00} & 79.85\scriptsize$\pm$0.33 & 3.55\scriptsize$\pm$0.39 \\
\textbf{Ours} & \textbf{93.71\scriptsize$\pm$0.01} & \textbf{0.00\scriptsize$\pm$0.00} & \textbf{90.96\scriptsize$\pm$0.02} & \textbf{0.00\scriptsize$\pm$0.00} \\
Upper Bound (\textit{UB}) & 95.6\scriptsize$\pm$0.01 & - & 93.56\scriptsize$\pm$0.01 & - \\
\bottomrule
\vspace{-20pt}
\end{tabular}
\end{adjustbox}
\caption{Results of domain-incremental learning on CORe50 \cite{core50}. `$\uparrow$' means higher is better and `$\downarrow$' means lower is better. Online-LoRA not only achieves the highest final accuracy but also demonstrates the lowest forgetting. }
\vspace{-20pt}
\label{table:dil}
\end{table}


\subsection{Exploration with length of task sequence} 
\label{sec:main-task_length}
\vspace{-3pt}

Table~\ref{table:taskSeq_disjoint} summarizes the results on Split ImageNet-S dataset across varying task sequence lengths; Table~\ref{table:taskSeq_blur} summarizes the results on Si-blurry ImageNet-S. As the task sequence is longer, all methods experience a decline in performance. However, Online-LoRA exhibits the smallest reduction in performance, showcasing its robustness against longer task sequences. This can be attributed to its utilization of loss surface plateaus, which effectively captures and adapts to shifts in data distribution at instance level. 

In contrast, for prompt-based learning methods such as L2P, longer task sequences challenge the capacity of prompt pool as more task-specific information needs to be encoded. Similarly, for replay-based methods, the strategy of selecting informative samples from the buffer is prone to biases in longer task sequences. This bias may result in an inadequate representation of earlier tasks or an overemphasis on more recent tasks, hurting the methods overall performance. 


\begin{table}[tb]
\centering
\begin{adjustbox}{max width=\linewidth}
\begin{tabular}{ccccc}
\toprule
Method & \multicolumn{2}{c}{10 tasks} & \multicolumn{2}{c}{20 tasks} \\
\cmidrule(lr){2-3} \cmidrule(lr){4-5}
 & $A_{\text{Final}}$ ($\uparrow$) & \textit{Forgetting} ($\downarrow$) & $A_{\text{Final}}$ ($\uparrow$) & \textit{Forgetting} ($\downarrow$) \\
\midrule
AGEM \cite{a-gem} & 0.16\scriptsize$\pm$0.04 & 9.42\scriptsize$\pm$0.17 & 0.11\scriptsize$\pm$0.05 & 7.96\scriptsize$\pm$0.10 \\
ER \cite{er} & 30.21\scriptsize$\pm$0.70 & 37.14\scriptsize$\pm$1.83 & 22.81\scriptsize$\pm$0.30 & 43.61\scriptsize$\pm$0.16 \\
EWC++ \cite{ewc++} & 0.32\scriptsize$\pm$0.28 & 22.46\scriptsize$\pm$4.69 & 0.11\scriptsize$\pm$0.05 & \textbf{5.26\scriptsize$\pm$0.45} \\
MIR \cite{mir} & 30.33\scriptsize$\pm$3.81 & 35.92\scriptsize$\pm$1.75 & 22.04\scriptsize$\pm$0.41 & 39.17\scriptsize$\pm$0.13 \\
GDumb \cite{gdumb} & 1.65\scriptsize$\pm$0.22 & - & 1.97\scriptsize$\pm$0.79 & - \\
PCR \cite{pcr} & 38.75\scriptsize$\pm$0.22 & 35.01\scriptsize$\pm$2.12 & 17.87\scriptsize$\pm$2.18 & 45.46\scriptsize$\pm$0.07 \\
DER++ \cite{der} & 6.47\scriptsize$\pm$0.06 & 15.34\scriptsize$\pm$0.15 & 2.29\scriptsize$\pm$0.23 & 23.14\scriptsize$\pm$0.06 \\
LODE (DER++) \cite{lode} & 9.97\scriptsize$\pm$2.29 & \textbf{8.48\scriptsize$\pm$1.24} & 13.47\scriptsize$\pm$0.66 & 35.89\scriptsize$\pm$1.63 \\
EMA (DER++) \cite{online-ema} & 16.88\scriptsize$\pm$2.23 & 36.28\scriptsize$\pm$1.09 & 11.55\scriptsize$\pm$0.66 & 38.56\scriptsize$\pm$0.22 \\
EMA (RAR) \cite{online-ema} & 14.06\scriptsize$\pm$0.37 & 36.28\scriptsize$\pm$1.09 & 9.05\scriptsize$\pm$0.60 & 29.77\scriptsize$\pm$1.70 \\
\textbf{Ours} & \textbf{47.06\scriptsize$\pm$0.24} & \textbf{28.09\scriptsize$\pm$3.25} & \textbf{44.19\scriptsize$\pm$2.09} & \textbf{28.48\scriptsize$\pm$0.24} \\
\midrule
Upper Bound (\textit{UB}) & \multicolumn{4}{c}{63.82\scriptsize$\pm$0.02} \\
\bottomrule
\vspace{-20pt}
\end{tabular}
\end{adjustbox}
\caption{Comparison with other methods on Split ImageNet-S for different lengths of task sequences. `$\uparrow$' means higher is better and `$\downarrow$' means lower is better. ViT-B/16 model is used.}
\label{table:taskSeq_disjoint}
\vspace{-5pt}
\end{table}



\begin{table}[tb]
\centering
\begin{adjustbox}{max height=0.3\linewidth}
\begin{tabular}{cccc}
\toprule
\multirow{2}{*}{Method} & \multirow{2}{*}{Task sequence} &
  \multicolumn{2}{c}{ImageNet-S} \\

  & & {$A_{\text{AUC}}$ ($\uparrow$)} & {$A_{\text{Final}}$ ($\uparrow$)} \\
  \midrule
L2P \cite{l2p} & \multirow{3}{*}{5 tasks} & 10.02\scriptsize$\pm$0.42 & 13.58\scriptsize$\pm$4.04 \\
MVP \cite{mvp} & & 10.68\scriptsize$\pm$0.45 & 13.99\scriptsize$\pm$1.73 \\
\textbf{Ours} & & \textbf{30.81\scriptsize$\pm$2.09} & \textbf{30.22\scriptsize$\pm$4.36}\\
\midrule
L2P \cite{l2p} & \multirow{3}{*}{10 tasks} & 9.06\scriptsize$\pm$0.43 & 12.49\scriptsize$\pm$3.39 \\
MVP \cite{mvp} & & 9.50\scriptsize$\pm$0.29 & 12.24\scriptsize$\pm$2.16 \\
\textbf{Ours} & & \textbf{30.69\scriptsize$\pm$0.59} & \textbf{31.44\scriptsize$\pm$4.39}\\
\midrule
L2P \cite{l2p} & \multirow{3}{*}{20 tasks} & 6.57\scriptsize$\pm$0.54 & 7.13\scriptsize$\pm$0.89 \\
MVP \cite{mvp} & & 7.87\scriptsize$\pm$0.24 & 8.98\scriptsize$\pm$1.49 \\
\textbf{Ours} & & \textbf{26.91\scriptsize$\pm$0.25} & \textbf{25.73\scriptsize$\pm$6.15}\\
\bottomrule
\vspace{-17pt}
\end{tabular}
\end{adjustbox}
\caption{Comparison with prompt-based methods on Si-blurry ImageNet-S at different length of task sequence. ViT-B/16 is used. }
\vspace{-15pt}
\label{table:taskSeq_blur}
\end{table}

\begin{figure*}[tb]
\centering
\begin{subfigure}{.33\textwidth}
  \centering
  \includegraphics[width=\linewidth]{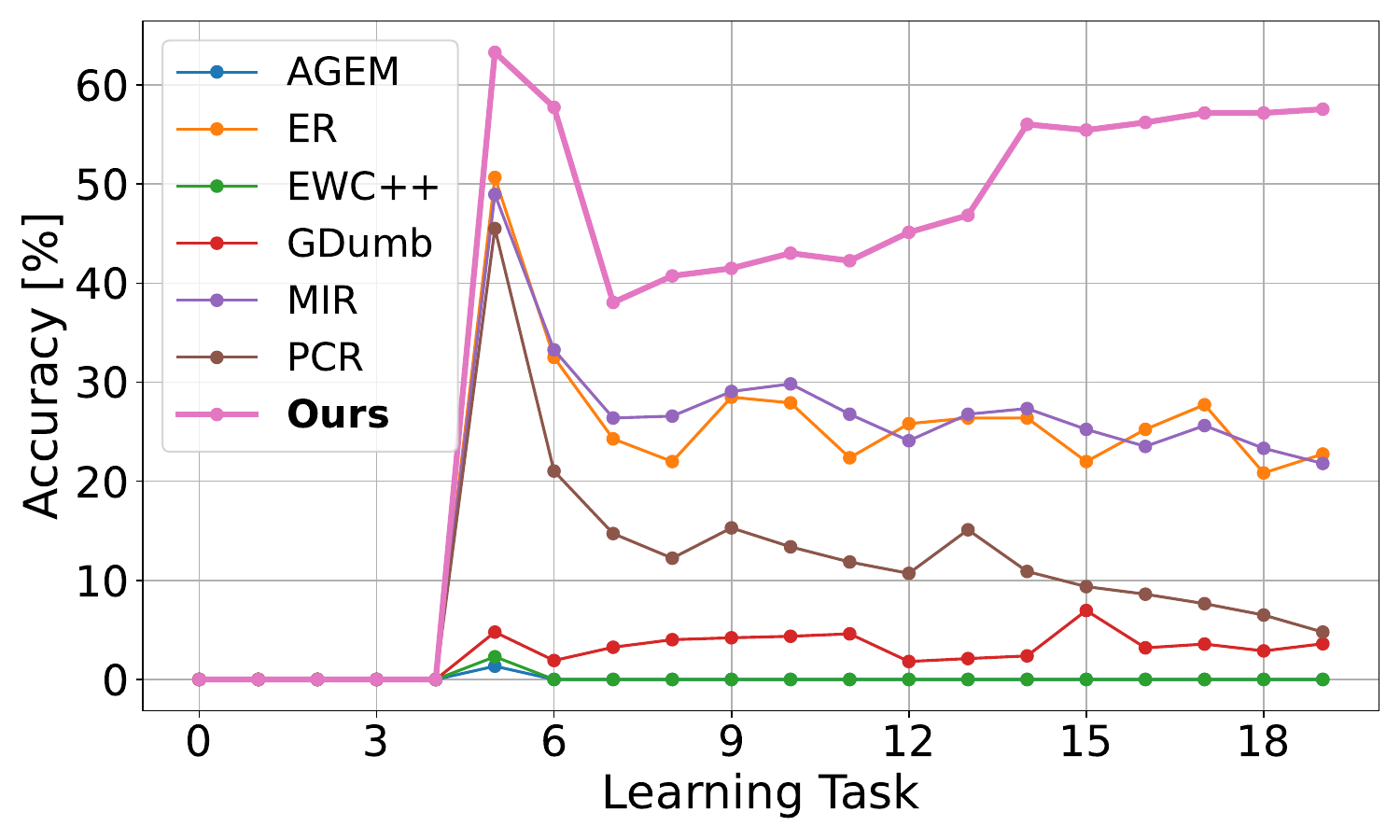}
  \caption{Task \#5}
  \label{fig:task5}
\end{subfigure}%
\begin{subfigure}{.33\textwidth}
  \centering
  \includegraphics[width=\linewidth]{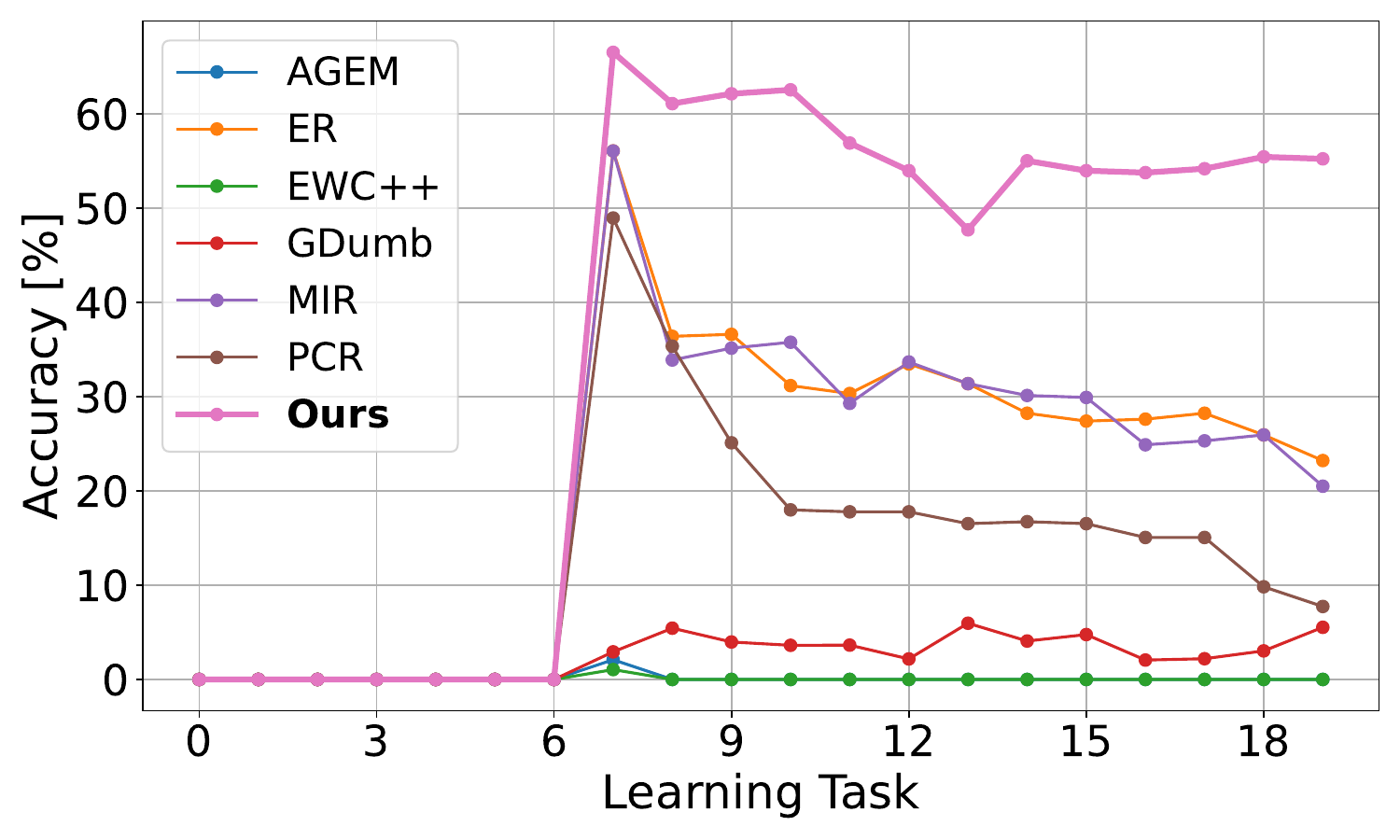}
  \caption{Task \#7}
  \label{fig:task7}
\end{subfigure}%
\begin{subfigure}{.33\textwidth}
  \centering
  \includegraphics[width=\linewidth]{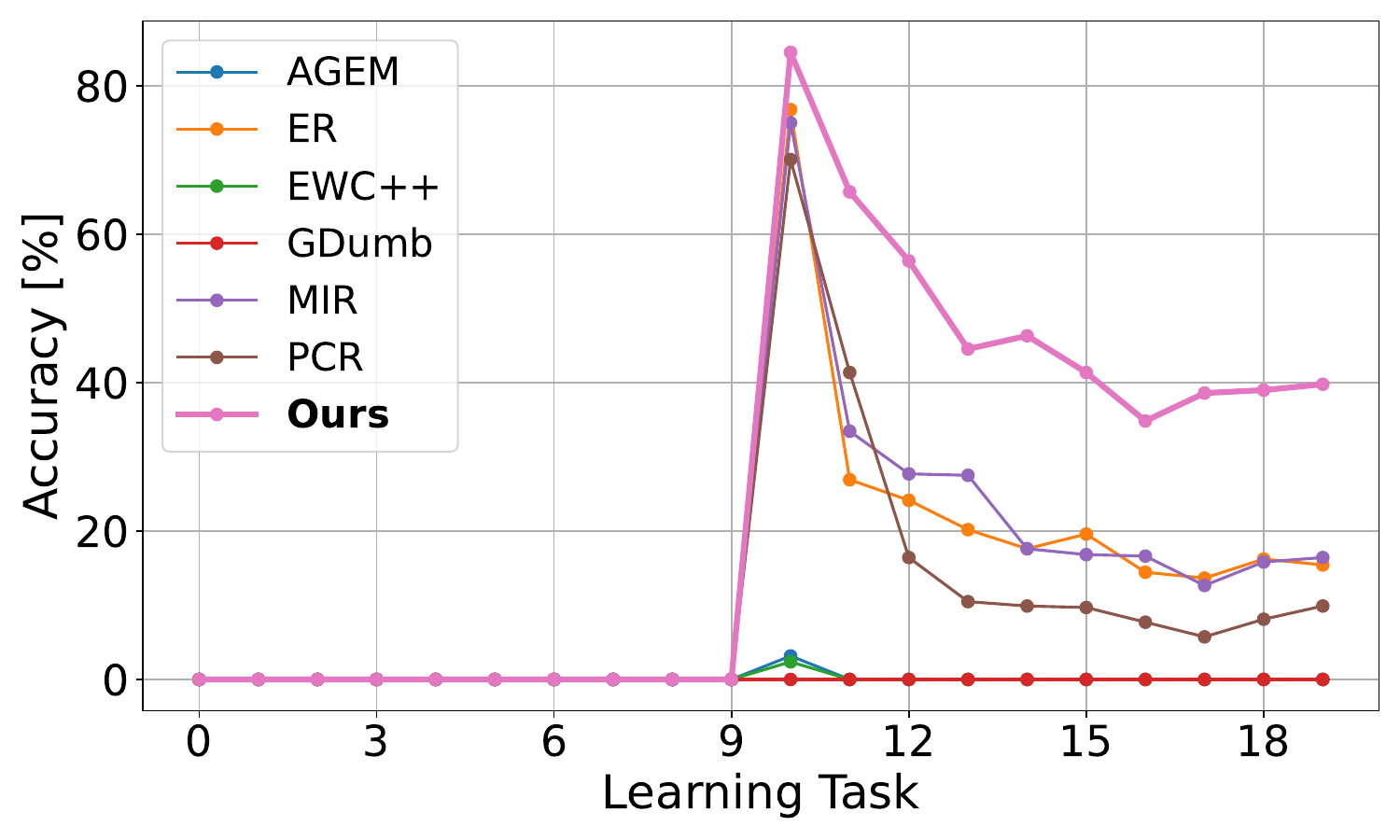}
  \caption{Task \#10}
  \label{fig:task10}
\end{subfigure}
\vspace{-10pt}
\caption{Test accuracy of three tasks versus the number of learning tasks. ViT-B/16 model is used on Split ImageNet-S with 20 tasks. The accuracy for each task prior to the model being trained on it is recorded as zero, since no measurements are taken at that stage, as the model has not yet been exposed to the corresponding task. 
}
\vspace{-10pt}
\label{fig:task accuracy_20tasks}
\end{figure*}


Furthermore, Figure~\ref{fig:task accuracy_20tasks} shows the accuracy on the validation set for four tasks at the time they are first encountered and after each subsequent task is learned (see Appendix~\ref{sec:supp-task-acc} for results of other tasks). As shown in Figure~\ref{fig:task accuracy_20tasks}, Online-LoRA consistently outperforms the other SOTA methods in terms of preserving the performance of previously learned tasks, which underscores the effectiveness of our online parameter regularization in mitigating catastrophic forgetting. 


\subsection{Ablation study}
\label{sec:main-ablation_study}
\vspace{-5pt}
Table~\ref{table:ablation} shows the ablation study on the effectiveness of each component (``incremental LoRA'' and ``hard loss'') of Online-LoRA on Split ImageNet-R (10 tasks). The results demonstrate the crucial role of each component of Online-LoRA in overall performance. More results in Appendix~\ref{supp-ablation}. 

Simply fine-tuning a single set of LoRA parameters (i.e. without incorporating any components of Online-LoRA) results in significantly worse performance compared to our approach, with a 20\% drop in accuracy (from 48.23\% to 28.68\%). Additionally, excluding the loss from hard buffer samples within the Online-LoRA framework leads to a substantial performance decline from 48.23\% to 34.74\% (a 13.5\% decrease). This emphasizes the crucial role of maintaining a minimal buffer with only the four most challenging samples in mitigating forgetting. 

Furthermore, the absence of new LoRA initialized at plateaus of the loss surface and model parameter regularization results in a significant performance decline of 12\%, from 48.23\% to 36.08\%. This highlights the importance of continuously adding new LoRA parameters to minimize task interference and implementing online weight regularization to prevent catastrophic forgetting. 




\begin{table}[tb]
\centering
\begin{adjustbox}{max width = \linewidth}
\begin{tabular}{cc|cc}
\toprule
Incremental LoRA & Hard loss & $A_\text{Final}$ ($\uparrow$) & $Forgetting$ ($\downarrow$)  \\
\midrule
- & - & 28.68\scriptsize$\pm$0.13 & 53.45\scriptsize$\pm$0.04\\
\checkmark & - & 34.74\scriptsize$\pm$0.31 & 34.37\scriptsize$\pm$1.15 \\
- & \checkmark & 36.08\scriptsize$\pm$0.19 & 35.75\scriptsize$\pm$0.33\\
\midrule
\checkmark & \checkmark & \textbf{48.23\scriptsize$\pm$0.74} & \textbf{23.85\scriptsize$\pm$1.08}\\
\bottomrule
\vspace{-20pt}
\end{tabular}
\end{adjustbox}
\caption{Ablation results of ViT-B/16 model on Split ImageNet-R dataset. `$\uparrow$' means higher is better and `$\downarrow$' means lower is better. "Incremental LoRA": introducing new, trainable LoRA at each loss plateau with the model parameter regularization in Equation~\ref{main-lora-loss}. . "Hard loss": including $\mathcal{L}(F(X_B; \theta), Y_B)$ (the loss from hard buffer samples) in the final learning objective in Equation~\ref{main-final-objective}. $\checkmark$ indicates the presence of the component, $-$ indicates its absence. }
\vspace{-15pt}
\label{table:ablation}
\end{table}

\vspace{-5pt}
\section{Conclusion}
\vspace{-5pt}
In this paper, we have presented Online-LoRA, a novel method for task-free online CL. Online-LoRA dynamically analyzes the loss surface to adapt the model to changing data distributions and uses online weight regularization to prevent catastrophic forgetting. 

We have also provided empirical evidence to show the effectiveness of Online-LoRA across various scenarios. Notably, Online-LoRA shows substantial performance advantage over other state-of-the-art methods in scenarios involving long task sequences. Furthermore, Online-LoRA's performance closely approaches the upper bound in domain-incremental settings. Given the widespread adoption of pre-trained models in CL, Online-LoRA offers a strong foundation for practical task-free online CL systems. 

\clearpage
{\small
\bibliographystyle{ieee_fullname}
\bibliography{egbib}
}

\clearpage

\appendix

In this supplementary materials, a unique labeling with an "S" prefix (e.g., S1, S2, etc.) is used, distinguishing them from the main paper references.

\section{Evaluation Metrics}
\label{sec:supp-metrics}

In this section, we present the definitions of the three evaluation metrics we used in our experiments, supplementing Section~\ref{main-exp-detail} in the main paper. 

Let $a_{i,j}$ be the testing accuracy on the $i^{th}$ task after training on $j^{th}$ task. The total number of tasks is denoted by $T$. 

\paragraph{Final Accuracy} The final accuracy $A_\text{Final}$ is calculated as the average accuracy across all tasks after training on the final task: 

\begin{equation}
    A_\text{Final} = \frac{1}{T} \sum^{T}_{i=1}a_{i, T}
\end{equation}

\paragraph{Area Under the Curve of Accuracy} The $A_\text{AUC}$ (Area Under the Curve of Accuracy) is defined as the area under the curve (AUC) of the accuracy-to-{\# of samples} curve \cite{auc}. To construct the curve, the accuracy is measure after each sample is observed. $A_\text{AUC}$ measures the any time inference accuracy of the model: 

\begin{equation}
A_{\text{AUC}} = \sum_{i=1}^{k} f(i \cdot \Delta n) \cdot \Delta n,
\end{equation}

where the step size $\Delta n$ is defined as $\Delta n = 1$, representing the number of samples observed between inference queries, and $f(\cdot)$ denotes the curve in the accuracy-to-\{number of samples\} plot. A high $A_{\text{AUC}}$ indicates that the method consistently maintains high accuracy throughout training.

\paragraph{Forgetting} Forgetting is defined as the averaged differences between the historical maximum accuracy of task $k$ and the accuracy of task $k$ after all tasks finish training: 

\begin{equation}
    \text{Forgetting} = \frac{1}{T-1} \sum^{T-1}_{k=1}\max_{t=1,2,...,T-1}(a_{k,t}-a_{k,T})
\end{equation}

The last task $T$ is excluded because the forgetting of the last task is always 0. 

\section{Experimental Details}
\label{sec:supp-exp_detail}

In this section, we provide details of the experiments we reported in the paper, supplementing Section~\ref{main-exps} in the main paper. 

\paragraph{Data preprocessing}
Because we focus on the ViT architectures ViT-B/16 and ViT-S/16, all input images are resized to 224$\times$224 and normalized to [0, 1]. 

\paragraph{Hyperparameters}

For tuning the threshold values for each dataset (CIFAR-100 \cite{cifar100}, ImageNet-R \cite{imagenet-r}, ImageNet-S \cite{imagenet-s}, CUB-200 \cite{cub-200}, and CORe50 \cite{core50}), we conducted a grid search following the protocol in \cite{ocl-survey}. The threshold grid is shown in Table~\ref{tab:supp-our-hyperparameter-tuning}. Table \ref{tab:supp-thresholds} shows the threshold values we used in our experiments. For CIFAR-100, ImageNet-R, and ImageNet-S, these threshold values remain consistent in both disjoint and Si-blurry class-incremental scenarios. 

We set the regularization factor $\lambda$=2000.0 (see Equation~\ref{main-lora-loss} in the main paper) for all experiments. 

\begin{table*}[tb]
\centering
\begin{adjustbox}{max width = \textwidth}
\begin{tabularx}{\textwidth}{lXXXXX}
\toprule
Threshold & CIFAR-100 & ImageNet-R & ImageNet-S & CUB-200 & CORe50 \\
\midrule
Mean & [2.2, 2.6, 2.8, 3.0] & \multicolumn{3}{c}{[5.2, 5.4, 5.6, 5.8, 6.0]} & [18.0, 24.0, 30.0] \\
Variance & \multicolumn{4}{c}{[0.02, 0.03, 0.04, 0.06, 0.08, 0.1]} & [0.6, 0.8, 1.0, 1.2] \\
\bottomrule
\end{tabularx}
\end{adjustbox}
\caption{Hyperparameter grid for the mean and variance threshold values of the loss window in our Online-LoRA. }
\label{tab:supp-our-hyperparameter-tuning}
\end{table*}

\begin{table}[ht]
\centering
\begin{adjustbox}{max width=\linewidth}
\begin{tabular}{cccccc}
\toprule
Threshold & CIFAR-100 & ImageNet-R & ImageNet-S & CORe50 & CUB-200\\
\midrule
Mean & 2.6 & 5.2 & 5.6 & 6.0 & 24.0\\
Variance & 0.03 & 0.02 & 0.06 & 0.1 & 1.0\\
\bottomrule
\end{tabular}
\end{adjustbox}
\caption{Mean and variance thresholds of the loss window for different datasets. }
\label{tab:supp-thresholds}
\end{table}

\section{Loss Surface}
\label{sec:supp-loss_surface}

\begin{figure}[tb]
  \centering
  \includegraphics[width=\linewidth]{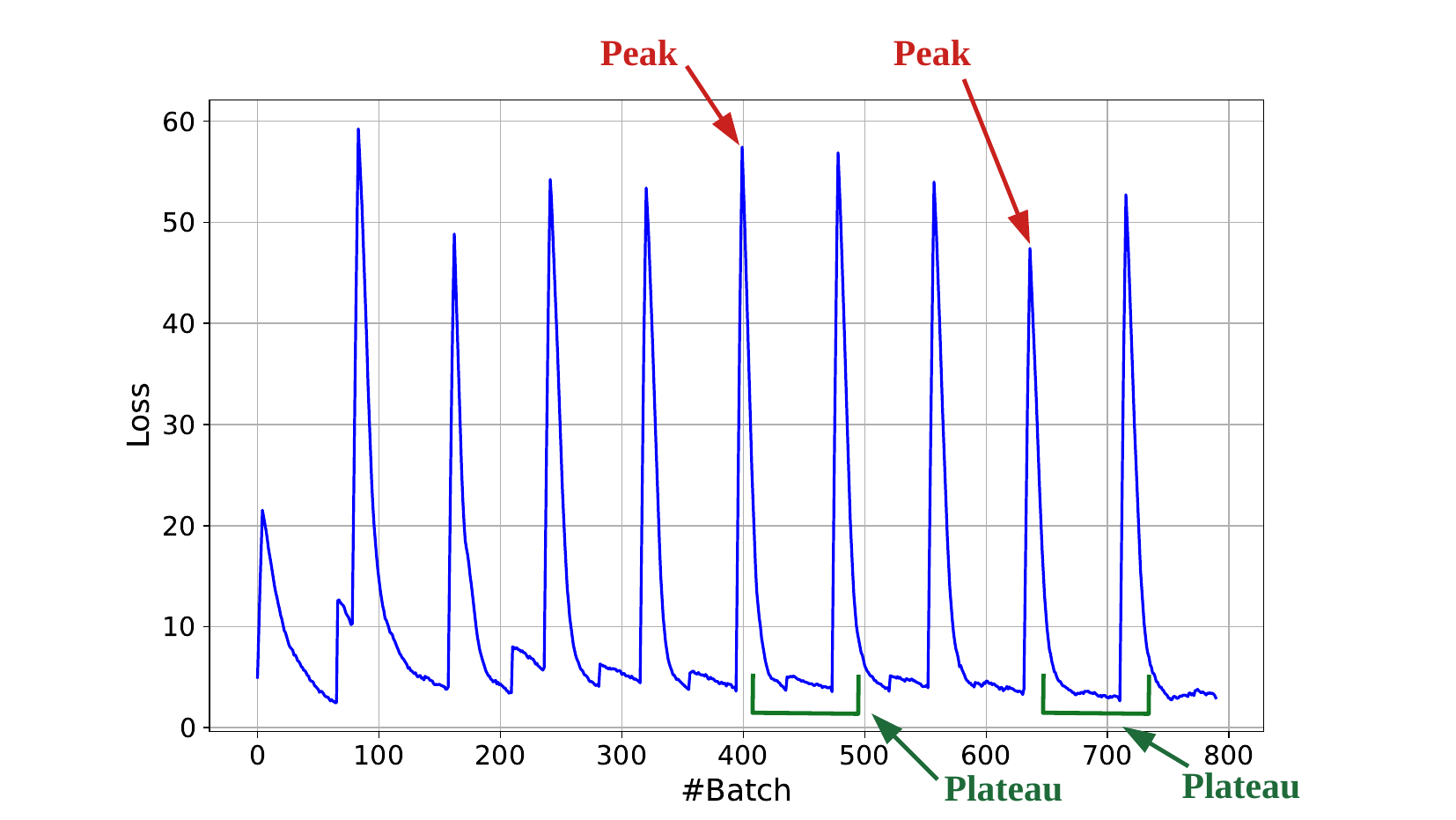}
  \caption{Loss surface of Online-LoRA on Split CIFAR-100 using ViT-B/16 model. Note that other peaks and plateaus exist but are not marked. }
  \label{fig:loss-surface-split}
\end{figure}

Figure~\ref{fig:loss-surface-split} shows more qualitative examples of how the loss surface recognizes data distribution shifts, supplementing Section~\ref{main-plateau} in the main paper. MAS \cite{mas_ocl} introduces the \textit{loss surface} to derive information about incoming streaming data in the task-free scenario. As shown in Figure~\ref{fig:loss-surface-split}, the peaks on the loss surface indicate shifts in the input data distribution. And the stable regions, namely plateaus, signal the convergence of the model. For instance, the Split CIFAR-100 dataset has 10 distinct tasks, with the data distribution remaining constant within each task. As a result, during the learning process of Split CIFAR-100, there are 9 shifts in data distribution, corresponding to 9 peaks in the loss surface, as illustrated in Figure \ref{fig:loss-surface-split}. 

To identify plateaus on the loss surface, we employ a \textit{loss window}, which is a sliding window that moves across consecutive training losses. Within this window, we closely observe both the mean and variance of the losses. A plateau is identified when both metrics fall below a predefined threshold (see Appendix~\ref{sec:supp-exp_detail} for details). Upon detecting a plateau, we proceed to introduce new LoRA parameters and update the estimation of the model parameter importance. Our goal in identifying plateaus is to mark periods of stable prediction following shifts in data distribution. Therefore, we only classify a phase as a plateau if it follows a peak. A peak is recognized when the loss window's mean increases by an amount exceeding the standard deviation of the window within a single batch. 

\section{Results of Swin Transformer}
\label{sec:supp-swin}

In this section, we present the results for the disjoint class-incremental and domain-incremental settings (for details on these settings, see Section~\ref{benchmark} in the main paper) using the Swin Transformer architecture \cite{swin}. For a fair comparison, the hyperparameters for the baseline methods are set according to the descriptions in Appendix~\ref{sec:supp-baseline-train-setting}. For our method, we use a learning rate of 0.0003 for the Swin Transformer.

As shown in Table~\ref{tab:supp-swin-result}, our approach consistently outperforms other baseline methods in both disjoint class-incremental and domain-incremental learning settings. This demonstrates that our method remains effective across various ViT architectures, extending beyond the ViT-B/16 and ViT-S/16 models reported in Section~\ref{sec:main-results} of the main paper.

\begin{table*}[tb]
\centering
\begin{adjustbox}{max width=\textwidth}
\setlength{\tabcolsep}{6pt}
\begin{tabular}{lcccc}
\toprule
\multirow{2}{*}{Method} & \multicolumn{2}{c}{Split-ImageNet-S} & \multicolumn{2}{c}{CORe50} \\
\cmidrule(lr){2-3} \cmidrule(lr){4-5}
 & $A_{\text{Final}}$ ($\uparrow$) & \textit{Forgetting} ($\downarrow$) & $A_{\text{Final}}$ ($\uparrow$) & \textit{Forgetting} ($\downarrow$) \\
\midrule
AGEM \cite{a-gem} & 31.67\scriptsize$\pm$0.96 & 50.12\scriptsize$\pm$0.27 & 90.15\scriptsize$\pm$1.31 & 1.16\scriptsize$\pm$0.05 \\
ER \cite{er} & 42.60\scriptsize$\pm$0.75 & 38.68\scriptsize$\pm$0.26 & 88.93\scriptsize$\pm$2.99 & 4.16\scriptsize$\pm$0.09 \\
EWC++ \cite{ewc++} & 29.57\scriptsize$\pm$1.57 & 51.87\scriptsize$\pm$0.04 & 90.91\scriptsize$\pm$1.28 & 0.04\scriptsize$\pm$0.02 \\
MIR \cite{mir} & 42.90\scriptsize$\pm$0.19 & 38.49\scriptsize$\pm$0.15 & 87.47\scriptsize$\pm$0.65 & 5.67\scriptsize$\pm$0.14 \\
GDumb \cite{gdumb} & 14.76\scriptsize$\pm$1.13 & - & 79.52\scriptsize$\pm$3.00 & - \\
\textbf{Ours} & \textbf{53.75\scriptsize$\pm$0.29} & \textbf{32.86\scriptsize$\pm$0.89} & \textbf{95.29\scriptsize$\pm$0.06} & \textbf{0.00\scriptsize$\pm$0.00} \\
\midrule
\textit{UB} & 71.98\scriptsize$\pm$0.23 & - & 97.56\scriptsize$\pm$0.02 & - \\
\bottomrule
\end{tabular}
\end{adjustbox}
\caption{Results of disjoint class-incremental learning and domain-incremental learning using Swin Transformer. `$\uparrow$' means higher is better and `$\downarrow$' means lower is better. The best results are noted by \textbf{bold}. \textit{UB} is the upper-bound performance. With Swin Transformer, our Online-LoRA method consistently outperforms other baseline methods across various settings, demonstrating its adaptability and effectiveness across different ViT architectures. }
\label{tab:supp-swin-result}
\end{table*}

\section{Supplementary Ablation Study}
\label{supp-ablation}

\subsection{Ablation Study on Imagenet-S Dataset}

In addition to the ablation results on Split Imagenet-R presented in Section~\ref{sec:main-ablation_study} of the main paper, this section provides further ablation results on the Split Imagenet-Sketch dataset with varying task lengths. As shown in Table~\ref{table:supp-ablation}, our Online-LoRA consistently outperforms other variants that lack certain components. These results demonstrate that both the hard buffer loss and incremental LoRA, along with online parameter regularization, are crucial for the performance of our approach. 

The baseline involves continuous fine-tuning of a single set of LoRA parameters. In contrast, Online-LoRA introduces an incremental LoRA architecture coupled with parameter importance-based regularization, and preserves a hard buffer along with its loss computations. Individually, each component improves performance and reduces forgetting. However, integrating both components into the baseline achieves the optimal performance, demonstrating the efficacy of our complete approach.

\begin{table*}[tb]
\centering
\begin{adjustbox}{max width = \textwidth}
\begin{tabular}{cc|cc|cc}
\toprule
\multirow{2}{*}{Incremental LoRA} & \multirow{2}{*}{Hard loss} & \multicolumn{2}{c|}{10 tasks} & \multicolumn{2}{c}{20 tasks} \\
\cmidrule{3-6}
 & & $A_\text{Final}$ ($\uparrow$) & $Forgetting$ ($\downarrow$) & $A_\text{Final}$ ($\uparrow$) & $Forgetting$ ($\downarrow$)  \\
\midrule
- & - & 30.66\scriptsize$\pm$0.25 & 38.70\scriptsize$\pm$0.40 & 24.49\scriptsize$\pm$2.61 & 39.29\scriptsize$\pm$2.57\\
\checkmark & - & 31.11\scriptsize$\pm$2.60 & 34.62\scriptsize$\pm$2.98 & 32.47\scriptsize$\pm$0.29 & 33.14\scriptsize$\pm$1.39 \\
- & \checkmark & 36.26\scriptsize$\pm$0.12 & 39.29\scriptsize$\pm$2.57 & 35.43\scriptsize$\pm$4.99 & 32.56\scriptsize$\pm$2.72\\
\midrule
\checkmark & \checkmark & \textbf{47.06\scriptsize$\pm$0.24} & \textbf{28.09\scriptsize$\pm$3.25} & \textbf{44.19\scriptsize$\pm$2.09} & \textbf{28.48\scriptsize$\pm$0.24}\\
\bottomrule
\end{tabular}
\end{adjustbox}
\caption{Ablation results of ViT-B/16 model on Split ImageNet-Sketch dataset. `$\uparrow$' means higher is better and `$\downarrow$' means lower is better. "Incremental LoRA": introducing new, trainable LoRA at each loss plateau with the model parameter regularization in Equation~\ref{main-lora-loss} in paper. "Hard loss": including $\mathcal{L}(F(X_B; \theta), Y_B)$ (the loss from hard buffer samples) in the final learning objective in Equation~\ref{main-final-objective} in paper. A check mark ($\checkmark$) indicates the presence of the component, while a dash (—) indicates its absence. }
\label{table:supp-ablation}
\end{table*}

\subsection{Impact of Pre-trained Weights} 
\label{sec: supp-justify-pretrained}

In this section, we demonstrate that our experimental settings do not provide any unfair advantage to our Online-LoRA approach through the use of pre-trained ViT models.

First, it is important to note that all baseline methods in our experiments utilize the same pre-trained ViT models as their backbones, just like Online-LoRA. Consequently, all methods benefit from the pre-training to varying extents, particularly those originally implemented with ResNet18 backbones (Table~\ref{tab:supp-res-vit-comp}). For detailed information on the backbones used by each baseline, please refer to Appendix~\ref{sec: supp-baseline-backbone}.

Second, we show that simply using pre-trained models without applying any CL methods or strategies fails to yield competitive performance. To illustrate this, we introduce three simple baselines:

\begin{itemize}
  \item \textbf{Frozen FT}: This baseline freezes the pre-trained backbone (feature extractor). Only the classification head (the final layer used for classification) is continuously fine-tuned on the data stream. Given that the model is pre-trained on the ImageNet-21K dataset, if any unfair advantage exists due to data leakage or other factors, it should be evident here by showing strong performance. 
  \item \textbf{Continual FT}: This baseline fully fine-tunes the pre-trained model, including both the backbone and the classification head, on each new data batch. This is consistent with our OCL setting where the model encounters each data batch only once. If the pre-trained weights alone brings any unfair advantage, this baseline should perform competitively, similar to methods specifically designed for CL. 
  \item \textbf{Random Head}: This baseline uses the pre-trained model's backbone with a newly initialized classifier head and performs only inference without any fine-tuning. Since the classifier head is randomly initialized, it should provide a clear lower bound for performance, demonstrating that without any adaptation or learning, the model's performance is essentially at chance level.
\end{itemize}

As shown in Table~\ref{tab:supp-pre-trained-online-ft}, \textbf{Random Head} baseline achieves near-zero accuracy, confirming that merely using pre-trained weights without adaptation to the test dataset does not have an advantage. Although the \textbf{Frozen FT} and \textbf{Continual FT} baselines outperform some CL methods (which also use the same pre-trained models), they still suffer from severe forgetting and exhibit a significant performance gap compared to other methods, particularly our Online-LoRA, with nearly a 20\% difference in final accuracy and a 30\% difference in forgetting.

These results demonstrate that the performance advantages of our Online-LoRA method over the baseline CL methods are not simply due to the use of pre-trained models. Instead, they arise from the effectiveness of our approach. The pre-trained weights provide a common foundation for all methods, but it is our approach that leads to superior performance.

\begin{table}[tb]
\centering
\begin{adjustbox}{max width = \linewidth}
\begin{tabular}{lcc}
\toprule
Method & Accuracy ($\uparrow$) & Forgetting ($\downarrow$)\\
\midrule
\midrule
Random Head & 0.08\scriptsize$\pm$0.00 & - \\
Frozen FT & 27.98\scriptsize$\pm$0.29 & 55.12\scriptsize$\pm$0.43 \\
Continual FT & 28.49\scriptsize$\pm$0.21 & 53.49\scriptsize$\pm$0.07 \\
\midrule
AGEM \cite{a-gem} & 5.60\scriptsize$\pm$2.74 & 53.97\scriptsize$\pm$1.97 \\
ER \cite{er} & 40.99\scriptsize$\pm$3.96 & 32.38\scriptsize$\pm$0.89\\
EWC++ \cite{ewc++} & 3.86\scriptsize$\pm$2.02 & 56.95\scriptsize$\pm$1.46\\
MIR \cite{mir} & 41.51\scriptsize$\pm$2.99 & 31.32\scriptsize$\pm$5.17\\
GDumb \cite{gdumb} & 1.65\scriptsize$\pm$0.22 & - \\
PCR \cite{pcr} & 46.11\scriptsize$\pm$3.03 & 25.50\scriptsize$\pm$0.41\\
DER++ \cite{der} & 30.90\scriptsize$\pm$8.04 & 24.26\scriptsize$\pm$4.14 \\
LODE (DER++) \cite{lode} & 42.20\scriptsize$\pm$6.46 & 31.83\scriptsize$\pm$1.05\\
EMA (DER++) \cite{online-ema} & 41.75\scriptsize$\pm$1.98 & 32.65\scriptsize$\pm$1.55 \\
EMA (RAR) \cite{online-ema} & 30.04\scriptsize$\pm$0.33 & 39.36\scriptsize$\pm$0.04 \\
\textbf{Online-LoRA (ours)} & \textbf{48.18\scriptsize$\pm$0.63} & \textbf{23.85\scriptsize$\pm$1.48}\\
\midrule
\textit{UB} & 63.82\scriptsize$\pm$0.02 & - \\
\bottomrule
\end{tabular}
\end{adjustbox}
\caption{Performance comparison between pre-trained models without CL strategies and pre-trained models with CL strategies on Split ImageNet-R (online class-incremental learning setting). ViT-B/16 backbone is used. While some methods do not outperform simple fine-tuning on a continuous data stream, other CL methods provide significant performance improvements to the pre-trained model. This demonstrates that the advantages of CL methods, including Online-LoRA, are not solely due to the use of pre-trained weights but also stem from the effectiveness of the methods themselves. UB is the upper-bound baseline trained on the i.i.d. data of the datasets. The best results are noted by \textbf{bold}. }
\label{tab:supp-pre-trained-online-ft}
\end{table}

\section{Baseline Settings}
\label{sec:supp-baseline-setting}

In this section, we provide the experimental settings for the baseline methods used in our experiments\footnote{Codebases used: \href{https://github.com/AlbinSou/online_ema.git}{https://github.com/AlbinSou/online\_ema.git}, \href{https://github.com/liangyanshuo/Loss-Decoupling-for-Task-Agnostic-Continual-Learning.git}{https://github.com/liangyanshuo/Loss-Decoupling-for-Task-Agnostic-Continual-Learning.git}, \href{https://github.com/FelixHuiweiLin/PCR.git}{https://github.com/FelixHuiweiLin/PCR.git}, \href{https://github.com/RaptorMai/online-continual-learning.git}{https://github.com/RaptorMai/online-continual-learning.git}}. 

\subsection{Overview of Baselines}

\begin{itemize}
    \item \textbf{AGEM} \cite{a-gem}: Averaged Gradient Episodic Memory, utilizes samples in the memory buffer to constrain parameter updates.
    \item \textbf{ER} \cite{er}: Experience replay, a rehearsal-based method with random sampling in memory retrieval and reservoir sampling in memory update.
    \item \textbf{EWC++} \cite{ewc++}: An online version of EWC \cite{ewc}, a regularization method that limits the update of parameters crucial to past tasks.
    \item \textbf{MIR} \cite{mir}: Maximally Interfered Retrieval, a rehearsal-based method that retrieves memory samples with loss increases given the estimated parameter update based on the current batch.
    \item \textbf{GDumb} \cite{gdumb}: Greedy Sampler and Dumb Learner, a strong baseline that greedily updates the memory buffer from the data stream with the constraint to keep a balanced class distribution.
    \item \textbf{PCR} \cite{pcr}: Proxy-based contrastive replay, a rehearsal-based method that replaces the samples for anchor with proxies in a contrastive-based loss.
    \item \textbf{DER++} \cite{der}: Dark Experience Replay++, a rehearsal-based method using knowledge distillation from past experiences.
    \item \textbf{LODE} \cite{lode}: Loss Decoupling, a rehearsal-based method that decouples the learning objectives of old and new tasks to minimize interference.
    \item \textbf{EMA} \cite{online-ema}: Exponential Moving Average, a model ensemble method that combines models from various training tasks.
    \item \textbf{L2P} \cite{l2p}: Learning to Prompt, a prompt-based method that prepends learnable prompts selected from a prompt pool to the embeddings of a pre-trained transformer.
    \item \textbf{MVP} \cite{mvp}: Mask and Visual Prompt tuning, a prompt-based method that uses instance-wise feature space masking.
\end{itemize}

\subsection{Backbone}
\label{sec: supp-baseline-backbone}

Among the baseline methods we compare, L2P \cite{l2p} and MVP \cite{mvp} originally reported results using a ViT-B/16 model \cite{vit} pre-trained on ImageNet21k, while the other baselines (AGEM \cite{a-gem}, ER \cite{er}, EWC++ \cite{ewc++}, MIR \cite{mir}, GDumb \cite{gdumb}, DER++ \cite{der}, PCR \cite{pcr}, LODE \cite{lode}, EMA \cite{online-ema}) reported results using a ResNet18 \cite{resnet18} architecture.

To ensure a fair comparison, we standardized our experimental setup by evaluating all baselines using the same pre-trained ViT model (ViT-B/16 and ViT-S/16). For methods originally implemented with ResNet18, we reimplemented them with ViT to match the experimental conditions of L2P and MVP. As shown in Table~\ref{tab:supp-res-vit-comp}, all methods perform better with the pre-trained ViT-B/16 than with ResNet18, supporting our argument that using a pre-trained ViT provides a more consistent and stronger baseline for performance comparisons. 

\begin{table*}[tb]
\centering
\begin{adjustbox}{max width = \textwidth}
\begin{tabular}{lccc}
\toprule
Method & Acc. w/ ResNet18 & Acc. w/ ViT-B/16 & Performance Gain (\%) \\
\midrule
AGEM \cite{a-gem} & 5.4\scriptsize$\pm$0.6 & 12.67\scriptsize$\pm$1.87 & 134.63 \\
ER \cite{er} & 14.5\scriptsize$\pm$0.8 & 44.85\scriptsize$\pm$1.83 & 209.31 \\
EWC++ \cite{ewc++} & 4.8\scriptsize$\pm$0.2 & 10.61\scriptsize$\pm$0.74 & 121.04 \\
MIR \cite{mir} & 14.8\scriptsize$\pm$0.7 & 48.36\scriptsize$\pm$3.11 & 226.76 \\
GDumb \cite{gdumb} & 24.8\scriptsize$\pm$0.7 & 41.00\scriptsize$\pm$19.97 & 65.32 \\
PCR \cite{pcr} & 21.8\scriptsize$\pm$0.9 & 48.48\scriptsize$\pm$0.15 & 122.39 \\
DER++ \cite{der} & 15.5\scriptsize$\pm$1.0 & 36.64\scriptsize$\pm$6.11 & 136.39 \\
LODE (DER++) \cite{lode} & 37.8\scriptsize$\pm$1.1 & 44.29\scriptsize$\pm$1.48 & 17.17 \\
EMA (DER++) \cite{online-ema} & 23.2\scriptsize$\pm$1.2 & 42.28\scriptsize$\pm$4.36 & 82.24 \\
EMA (RAR) \cite{online-ema} & 35.4\scriptsize$\pm$1.2 & 47.10\scriptsize$\pm$0.82 & 33.05 \\
\bottomrule
\end{tabular}
\end{adjustbox}
\caption{Performance comparison on CIFAR-100 between ResNet18 and pre-trained ViT-B/16 in an online class-incremental learning scenario. Acc. stands for Accuracy. All rehearsal-based methods use a buffer size of 500 for fair comparison. The results demonstrate that there is no unfair comparison in our experiments, as all methods benefit from the pre-trained ViT-B/16 model. The performance gain is computed as the percentage increase from the ResNet18 accuracy to the ViT-B/16 accuracy for each method.}
\label{tab:supp-res-vit-comp}
\end{table*}

\subsection{Training Settings}
\label{sec:supp-baseline-train-setting}

The following settings are shared by the baseline methods (and our Online-LoRA) in the experiments:

\begin{itemize} 
  \item Buffer Size: 500. Methods using a buffer include AGEM \cite{a-gem}, ER \cite{er}, MIR \cite{mir}, GDumb \cite{gdumb}, PCR \cite{pcr}, DER++ \cite{der}, LODE \cite{lode}, and EMA \cite{online-ema}. 
  \item Optimizer: Adam. 
  \item Batch Size: 64. 
\end{itemize}

In Table~\ref{tab:supp-baseline-hyperparameter}, we summarize the hyperparameters used for all baseline methods in our experiments. To ensure a fair comparison, we adopted these hyperparameters from their original codebases. However, because the baseline methods used different backbones and batch sizes in their original experiments, we adjusted the learning rates for some baselines to standardize the comparison across all methods. For tuning the learning rates, we followed the protocol outlined in \cite{ocl-survey} and conducted a grid search on a small cross-validation set. The hyperparameter grid for the baselines is detailed in Table~\ref{tab:supp-hyperparameter-grid}. 

\begin{table*}[tb]
\centering
\begin{adjustbox}{max width = \textwidth}
\begin{tabularx}{\textwidth}{lXXXXX}
\toprule
Method & CIFAR-100 & ImageNet-R & ImageNet-S & CUB-200 & CORe50 \\
\midrule
\midrule
AGEM \cite{a-gem} & \multicolumn{5}{c}{LR=0.0001, WD=0.0001} \\
\midrule
ER \cite{er} & \multicolumn{5}{c}{LR=0.0001, WD=0.0001, Episode memory per batch=10} \\
\midrule
\multirow{2}{*}{EWC++ \cite{ewc++}} & \multicolumn{5}{c}{LR=0.0001, WD=0.0001, $\lambda$=100, $\alpha$=0.9} \\
& \multicolumn{5}{c}{Number of training batches after which the Fisher information will be updated: 50} \\
\midrule
MIR \cite{mir} & \multicolumn{5}{c}{LR=0.0001, WD=0.0001, Number of subsample=50} \\
\midrule
\multirow{2}{*}{GDumb \cite{gdumb}} & \multicolumn{5}{c}{LR=0.001, WD=0.0001, Minimal learning rate: 0.0005, } \\
& \multicolumn{5}{c}{Gradient clipping=10, Epochs to train for memory=30} \\
\midrule
\multirow{2}{*}{PCR} \cite{pcr} & \multicolumn{5}{c}{LR=0.0001, WD=0.0001, Episode memory per batch=10, } \\
& \multicolumn{5}{c}{Temperature=0.09, Warmup of buffer before retrieve=4} \\
\midrule
DER++ \cite{der} & \multicolumn{5}{c}{LR=0.0003, $\alpha$=0.2, $\beta$=0.5} \\
\midrule
LODE \cite{lode} & \multicolumn{5}{c}{LR=0.0003, $C$=1.0, $\rho$=0.1} \\
\midrule
EMA \cite{online-ema} & \multicolumn{5}{c}{LR=0.0002, $\lambda$ for warm-up: 0.9, $\lambda$=0.99} \\
\midrule
L2P \cite{l2p} & \multicolumn{5}{c}{LR=0.003, Size of the prompt pool=10, Length of a single prompt=10, Number of prepended prompt=4} \\
\midrule
MVP \cite{mvp} & \multicolumn{5}{c}{LR=0.005, $\gamma$=2.0, $m$=0.5, $\alpha$=0.5} \\
\bottomrule
\end{tabularx}
\end{adjustbox}
\caption{Hyperparameters for the baseline methods on ViT-B/16. LR: learning rate. WD: weight decay. }
\label{tab:supp-baseline-hyperparameter}
\end{table*}

\begin{table*}[tb]
\centering
\begin{adjustbox}{max width = \textwidth}
\begin{tabularx}{\textwidth}{lXXXXX}
\toprule
Method & CIFAR-100 & ImageNet-R & ImageNet-S & CUB-200 & CORe50 \\
\midrule
\midrule
\multirow{2}{*}{AGEM \cite{a-gem}} & \multicolumn{5}{c}{LR: [0.0001, 0.0003, 0.001, 0.003, 0.01, 0.03, 0.1]} \\
& \multicolumn{5}{c}{WD: [0.0001, 0.001, 0.01, 0.1]} \\
\midrule
\multirow{2}{*}{ER \cite{er}} & \multicolumn{5}{c}{LR: [0.0001, 0.0003, 0.001, 0.003]} \\
& \multicolumn{5}{c}{WD: [0.0001, 0.001, 0.01, 0.1]} \\
\midrule
\multirow{2}{*}{EWC++ \cite{ewc++}} & \multicolumn{5}{c}{LR: [0.0001, 0.001, 0.01, 0.1]} \\
& \multicolumn{5}{c}{WD: [0.0001, 0.001]} \\
\midrule
\multirow{2}{*}{MIR \cite{mir}} & \multicolumn{5}{c}{LR: [0.0001, 0.001, 0.01, 0.1]} \\
& \multicolumn{5}{c}{WD: [0.0001, 0.001]} \\
\midrule
\multirow{2}{*}{GDumb \cite{gdumb}} & \multicolumn{5}{c}{LR: [0.001, 0.01, 0.1} \\
& \multicolumn{5}{c}{WD: [0.0001, 0.000001]} \\
\midrule
\multirow{2}{*}{PCR} \cite{pcr} & \multicolumn{5}{c}{LR: [0.0001, 0.001, 0.01, 0.1]} \\
& \multicolumn{5}{c}{WD: [0.0001, 0.001]} \\
\midrule
DER++ \cite{der} & \multicolumn{5}{c}{LR: [0.0003, 0.003, 0.03]} \\
\midrule
LODE \cite{lode} & \multicolumn{5}{c}{LR: [0.0003, 0.003, 0.03]} \\
\midrule
EMA \cite{online-ema} & \multicolumn{5}{c}{LR: [0.0001, 0.0002, 0.0003, 0.0004, 0.0005]} \\
\midrule
\bottomrule
\end{tabularx}
\end{adjustbox}
\caption{Hyperparameter grid for the baseline methods using the ViT-B/16 backbone. LR: learning rate; WD: weight decay. Since L2P \cite{l2p} and MVP \cite{mvp} use the same backbone and batch size as in our experiments, their learning rates were not adjusted. }
\label{tab:supp-hyperparameter-grid}
\end{table*}

\section{Exploration with Buffer Size}
\label{sec:supp-buffer-size}

Table~\ref{table:supp-buff-size} we show more results of the impact of buffer sizes on the performance of replay-based methods (AGEM \cite{a-gem}, ER \cite{er}, GDumb \cite{gdumb}, MIR \cite{mir}). 

As shown in Table \ref{table:supp-buff-size}, when the buffer size increases, all replay-based methods see improvements in their performance across the benchmarks. Notably, when the buffer size hits 5000 (a large capacity; 20\% of the ImageNet-R training set, 12.5\% of the ImageNet-S training set), the difference in performance between GDumb and other replay-based methods narrows. This suggests that the sophisticated memory retrieval strategies employed by these other methods do not significantly outperform GDumb's simple approach of training directly on the buffered data. Moreover, the performance of rehearsal-based methods drops when the buffer size shrinks. This highlights the efficiency of our Online-LoRA, which achieves high performance using just a minimal buffer size of 4. 

\begin{table*}[tb]
\centering
\begin{adjustbox}{max width=\textwidth}
\setlength{\tabcolsep}{6pt}
\begin{tabular}{lccccc}
\toprule
Buffer size & Method & Split-ImageNet-R & Split-ImageNet-S & Core50\\
\midrule
\multirow{4}{*}{500} & AGEM \cite{a-gem} & 5.60\scriptsize$\pm$2.74 & 0.16\scriptsize$\pm$0.04 & 80.15\scriptsize$\pm$2.97 \\
 & ER \cite{er} & 40.99\scriptsize$\pm$3.96 & 30.21\scriptsize$\pm$0.70 & 85.85\scriptsize$\pm$1.35\\
 & MIR \cite{mir} & 41.51\scriptsize$\pm$2.99 & 30.33\scriptsize$\pm$3.81 & 74.35\scriptsize$\pm$4.07 \\
 & GDumb \cite{gdumb} & 8.87\scriptsize$\pm$1.36 & 1.65\scriptsize$\pm$0.22 & 77.20\scriptsize$\pm$3.49 \\
\midrule
\multirow{4}{*}{1000} & AGEM \cite{a-gem} & 7.16\scriptsize$\pm$1.56 & 0.23\scriptsize$\pm$0.04 & 78.73\scriptsize$\pm$3.87\\
 & ER \cite{er} &44.71\scriptsize$\pm$2.63 & 34.32\scriptsize$\pm$0.53 & 84.27\scriptsize$\pm$4.11\\
 & MIR \cite{mir} & 46.65\scriptsize$\pm$5.63 & 33.99\scriptsize$\pm$1.72 & 82.64\scriptsize$\pm$1.12\\
 & GDumb \cite{gdumb} & 19.19\scriptsize$\pm$1.36 & 2.71\scriptsize$\pm$0.12 & 78.09\scriptsize$\pm$3.75\\
\midrule
\multirow{4}{*}{5000} & AGEM \cite{a-gem} & 7.21\scriptsize$\pm$0.34 & 0.12\scriptsize$\pm$0.02 & 77.57\scriptsize$\pm$3.56\\
 & ER \cite{er} & 47.23\scriptsize$\pm$2.71 & 37.65\scriptsize$\pm$0.23 & 81.32\scriptsize$\pm$2.19\\
 & MIR \cite{mir} & \textbf{49.33\scriptsize$\pm$3.49} & 35.90\scriptsize$\pm$2.35 & 81.18\scriptsize$\pm$3.20\\
 & GDumb \cite{gdumb} & 46.08\scriptsize$\pm$0.64 & 9.68\scriptsize$\pm$0.28 & 69.42\scriptsize$\pm$1.06\\
 \midrule
 & \textbf{Ours} & 48.18\scriptsize$\pm$0.63 & \textbf{47.06\scriptsize$\pm$0.24} & \textbf{93.71\scriptsize$\pm$0.01}\\
 & \textit{UB} & 76.78\scriptsize$\pm$0.44 & 63.82\scriptsize$\pm$0.02 & 95.60\scriptsize$\pm$0.01\\
\bottomrule
\end{tabular}
\end{adjustbox}
\caption{Results of replay-based methods with different buffer size. $A_{\text{Final}}$ metric and ViT-B/16 model is used. Each dataset has 10 disjoint tasks. \textit{UB} is the upper-bound baseline trained on the i.i.d. data of the datasets. The best results are noted by \textbf{bold}. }
\label{table:supp-buff-size}
\end{table*}

\begin{table*}[tb]
\centering
\begin{adjustbox}{max width = \textwidth}
\begin{tabular}{lccccc}
\toprule
Method & \#params (M) & FLOPs ($\times 10^{15}$) & Training time ($s$) \\
\midrule
\midrule
AGEM \cite{a-gem} & 85.88 & 140.52 & 828.39 \\
ER \cite{er} & 85.88 & 140.05 & 849.43 \\
EWC++ \cite{ewc++} & 85.88 & 214.36 & 1076.53 \\
GDumb \cite{gdumb} & 85.88 & 18.44 & 2078.59 \\
MIR \cite{mir} & 85.88 & 161.04 & 1069.29 \\
\textbf{Ours} & 86.47 & 151.20 & 864.60 \\
\bottomrule
\end{tabular}
\end{adjustbox}
\caption{Computational statistics for Online-LoRA and baseline methods on CIFAR-100 in the online class-incremental learning scenario using the ViT-B/16 backbone. FLOPs are measured as 'forward FLOPs per GPU' using the DeepSpeed FLOPS Profiler \cite{deepspeed-github}. All experiments are conducted on a single NVIDIA A100 GPU. }
\label{tab:supp-flops}
\end{table*}

\section{Computation Analysis}
\label{sec: supp-computation}

In this section, we present the model parameter size, training FLOPs, and training time for our Online-LoRA and the baseline methods. 

As shown in Table~\ref{sec: supp-computation}, our Online-LoRA model introduces approximately 0.6M additional parameters due to the inclusion of LoRA parameters, which represents a negligible increase (0.69\%) compared to the original size of the ViT-B/16 model. Notably, our memory buffer contains only 4 data samples, whereas other baselines (except EWC++) require at least 500 samples in their buffers to achieve comparable performance (see Appendix~\ref{sec:supp-buffer-size} for more details). Regarding computational consumption measured by FLOPs during training, Online-LoRA demonstrates advantages over EWC++ \cite{ewc++}, thanks to our efficient computation of the importance weight matrix, as explained in Section~\ref{regularization} of the main paper. The extremely low FLOPs of GDumb \cite{gdumb} can be attributed to its design, which involves greedily updating the memory buffer without employing additional strategies. However, its training time is relatively high because retraining is triggered frequently to maintain a balanced memory buffer, which adds overhead despite the low FLOPs.

\section{Task Accuracy}
\label{sec:supp-task-acc}

\begin{figure*}[tb]
\centering
\begin{subfigure}{.5\textwidth}
  \centering
  \includegraphics[width=.9\linewidth]{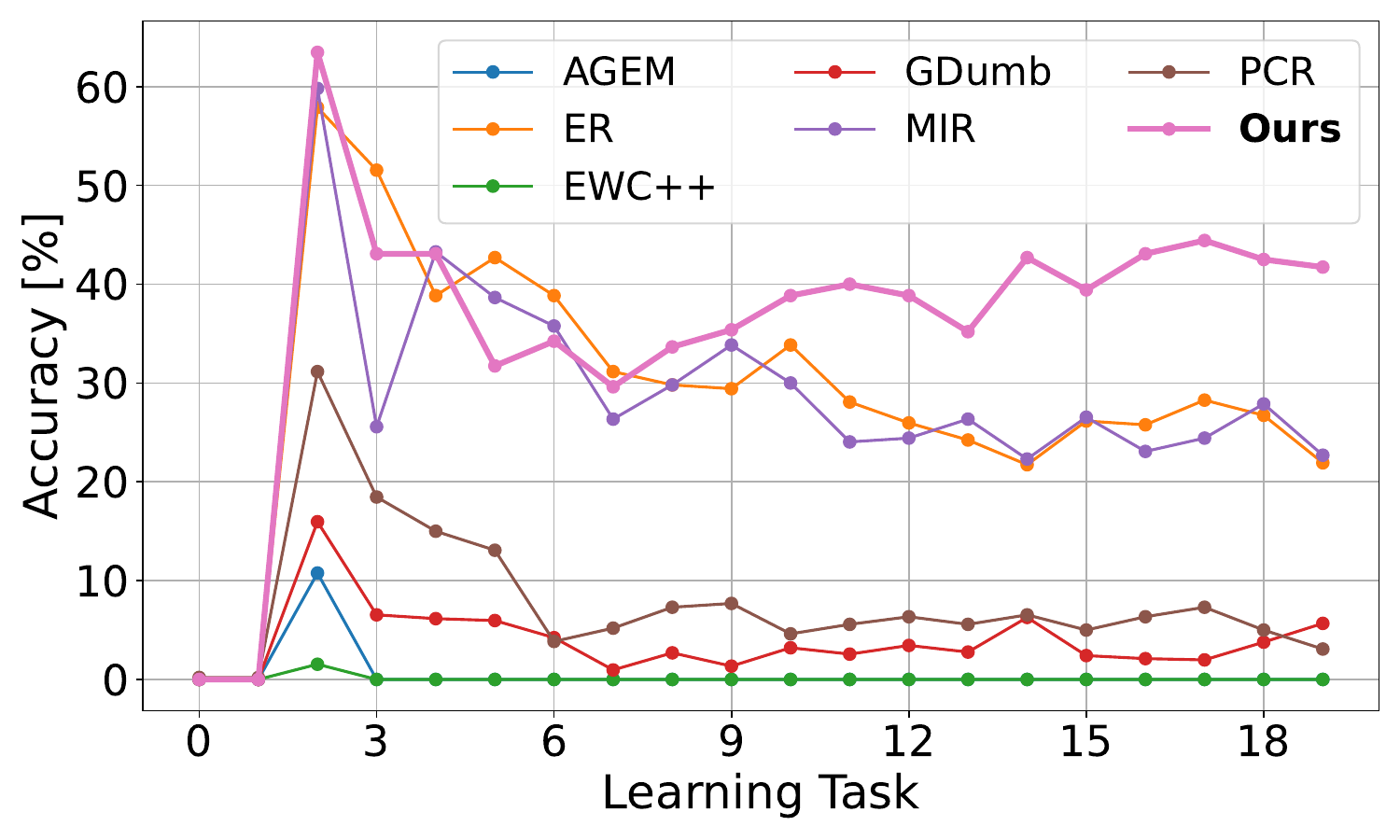}
  \caption{Task accuracy of task \#2}
  \label{fig:supp-task2}
\end{subfigure}
\begin{subfigure}{.5\textwidth}
  \centering
  \includegraphics[width=.9\linewidth]{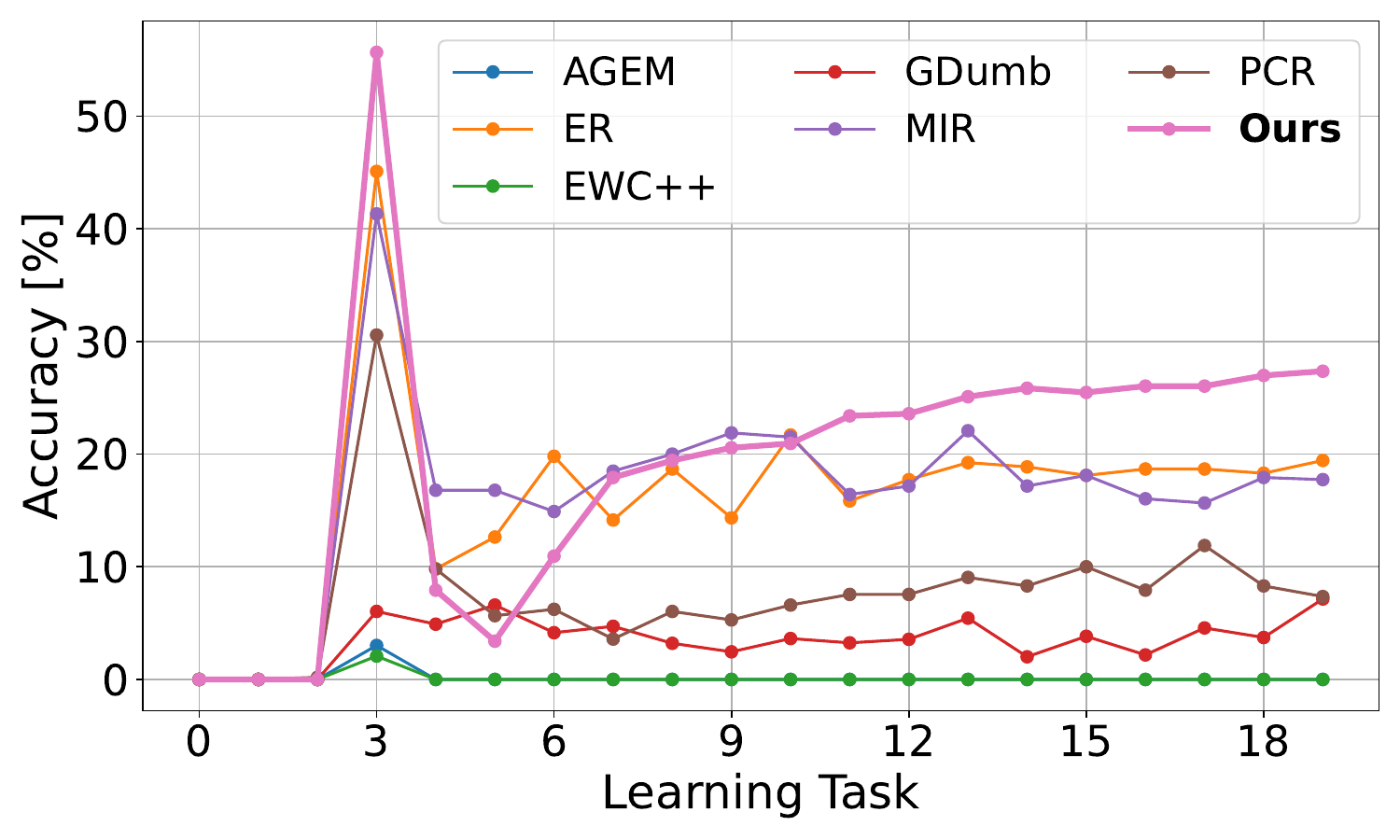}
  \caption{Task accuracy of task \#3}
  \label{fig:supp-task3}
\end{subfigure}
\begin{subfigure}{.5\textwidth}
  \centering
  \includegraphics[width=.9\linewidth]{plots/split_sketch_base_20tasktask5.pdf}
  \caption{Task accuracy of task \#5}
  \label{fig:supp-task5}
\end{subfigure}%
\begin{subfigure}{.5\textwidth}
  \centering
  \includegraphics[width=.9\linewidth]{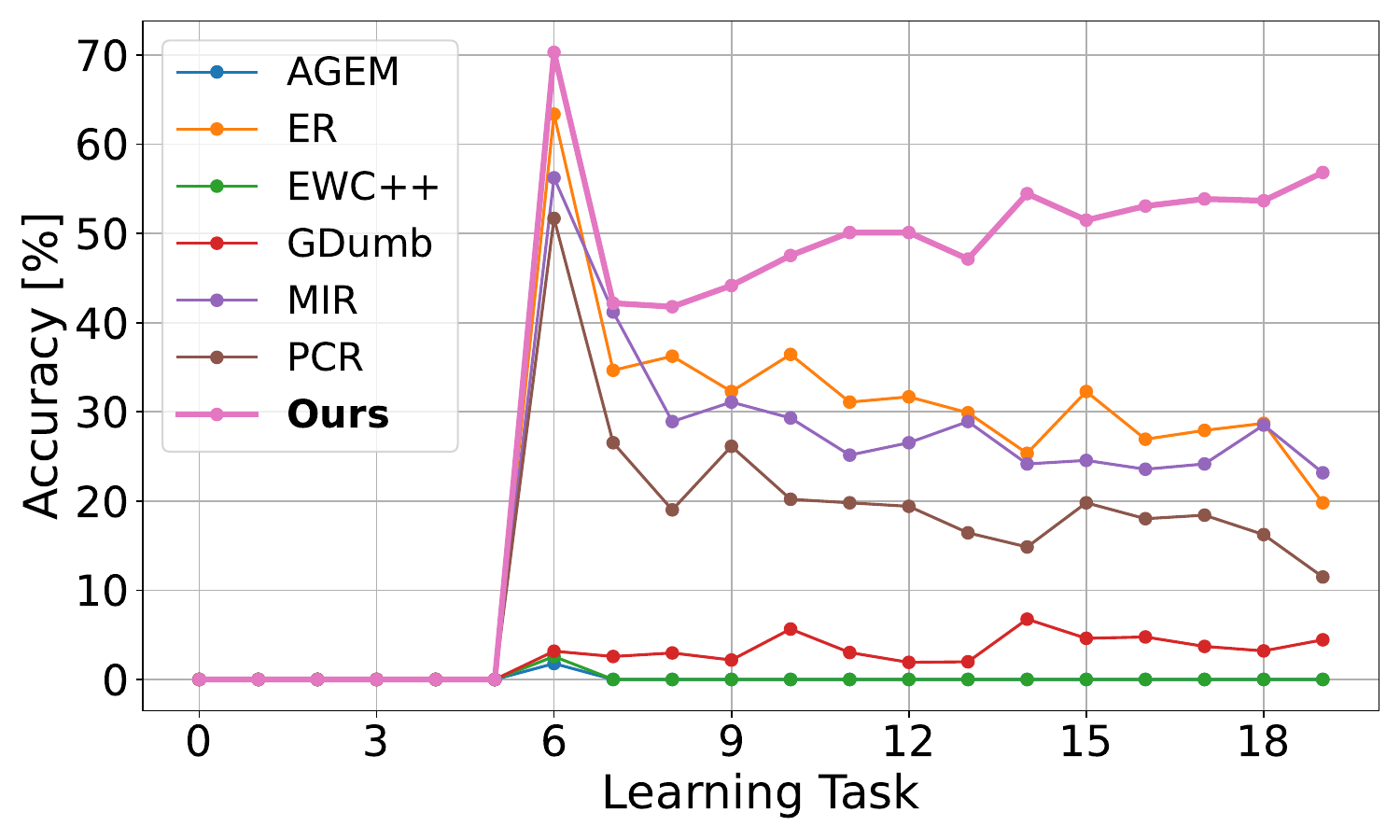}
  \caption{Task accuracy of task \#6}
  \label{fig:supp-task6}
\end{subfigure}
\begin{subfigure}{.5\textwidth}
  \centering
  \includegraphics[width=.9\linewidth]{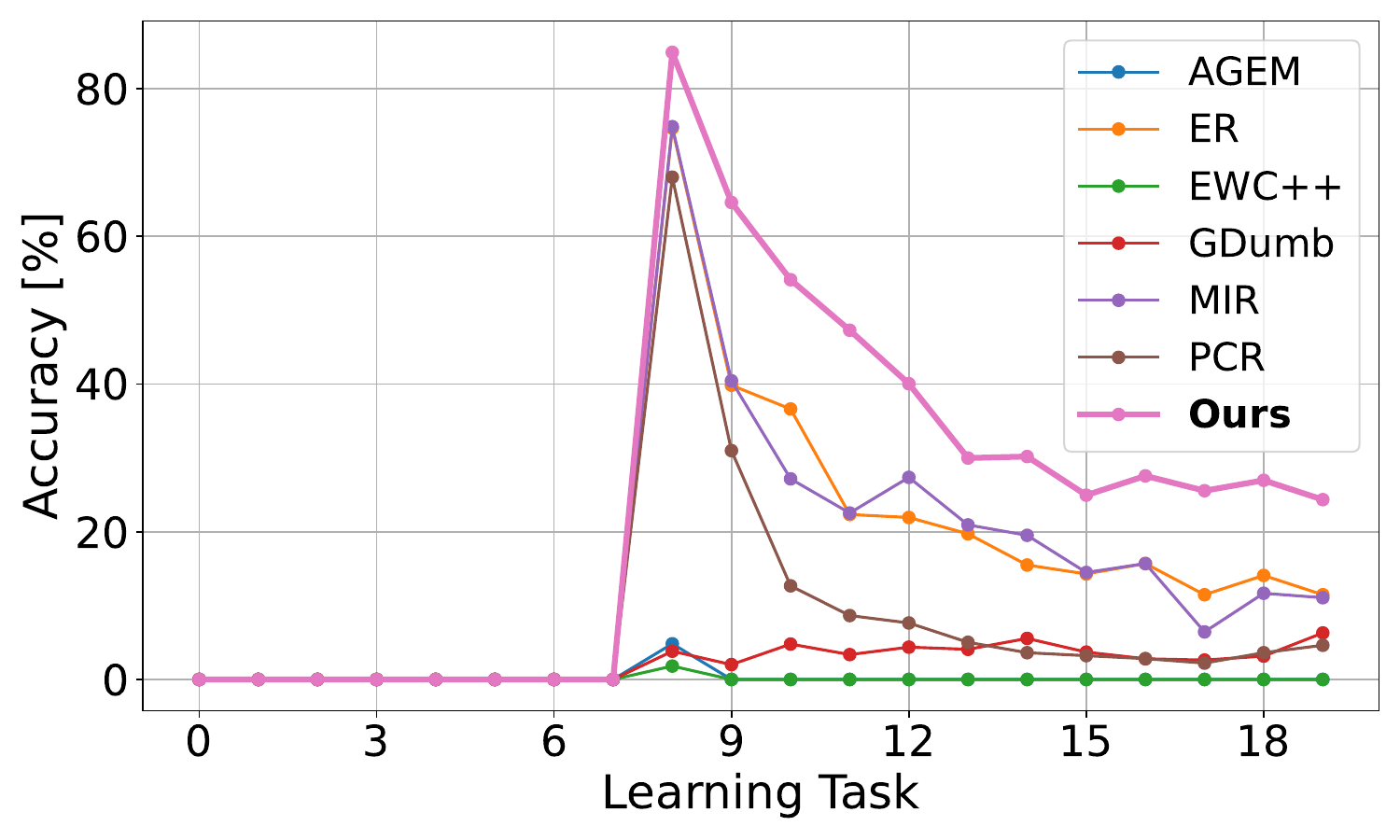}
  \caption{Task accuracy of task \#8}
  \label{fig:supp-task8}
\end{subfigure}%
\begin{subfigure}{.5\textwidth}
  \centering
  \includegraphics[width=.9\linewidth]{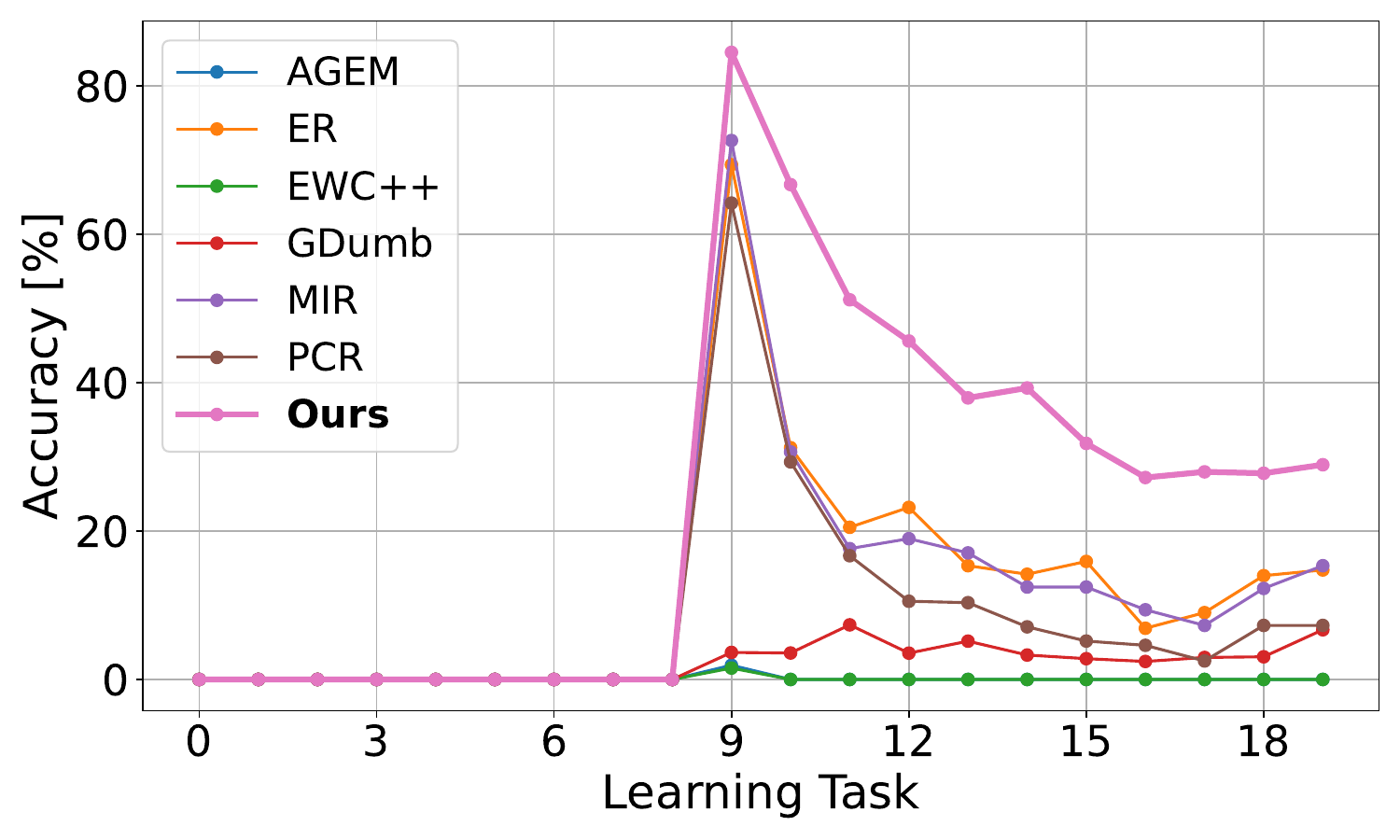}
  \caption{Task accuracy of task \#9}
  \label{fig:supp-task9}
\end{subfigure}
\caption{Task accuracy versus the number of learning tasks of task \#2 to task \#9. Our Online-LoRA consistently outperforms all the other methods in maintaining accuracy on previously learned tasks. Note that the recorded accuracy for initial tasks is zero, not due to poor model performance, but because our evaluation prioritizes mitigating forgetting in tasks the model has already encountered. }
\label{fig:supp-task-accuracy_2-9}
\end{figure*}

\begin{figure*}[tb]
\centering
\begin{subfigure}{.5\textwidth}
  \centering
  \includegraphics[width=.9\linewidth]{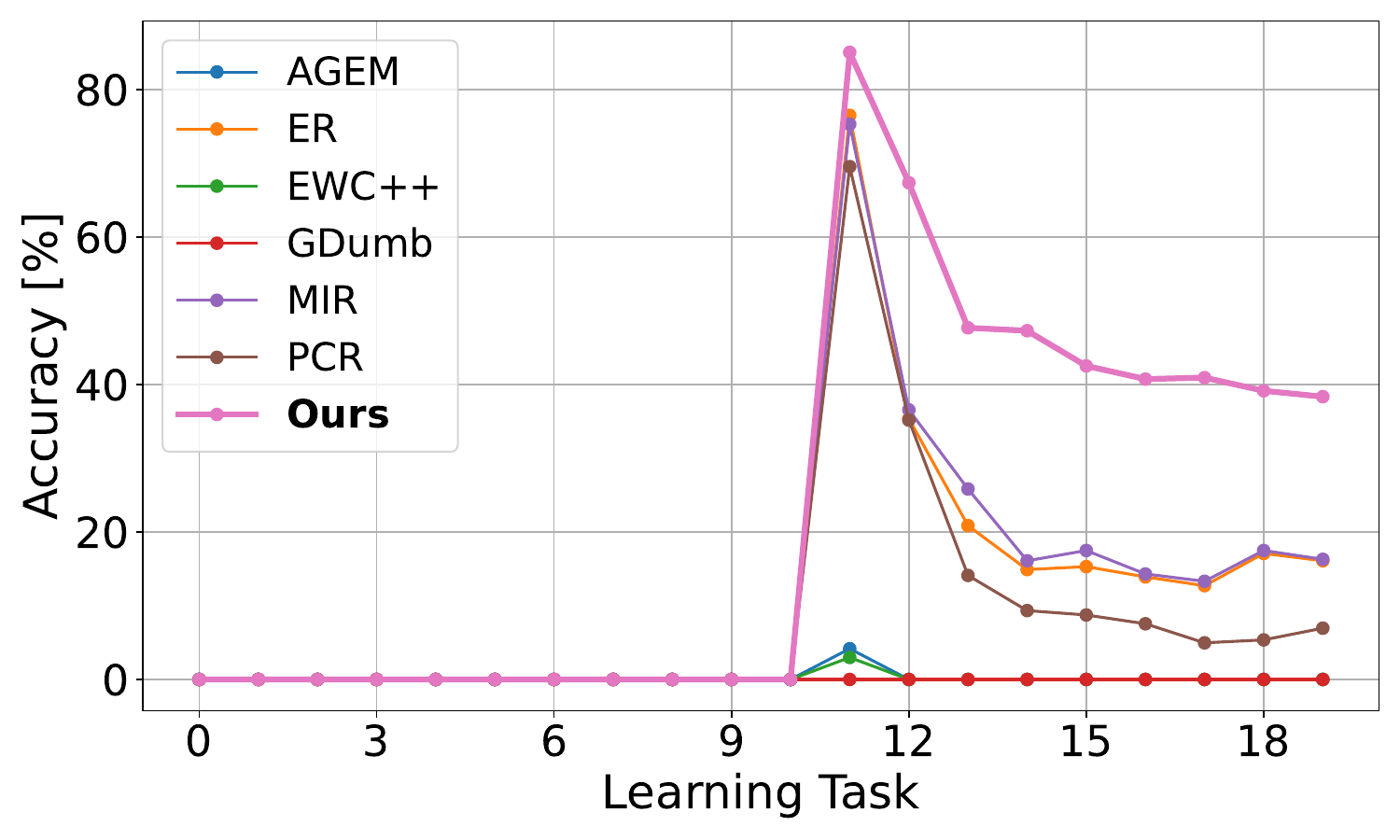}
  \caption{Task accuracy of task \#11}
  \label{fig:supp-task11}
\end{subfigure}%
\begin{subfigure}{.5\textwidth}
  \centering
  \includegraphics[width=.9\linewidth]{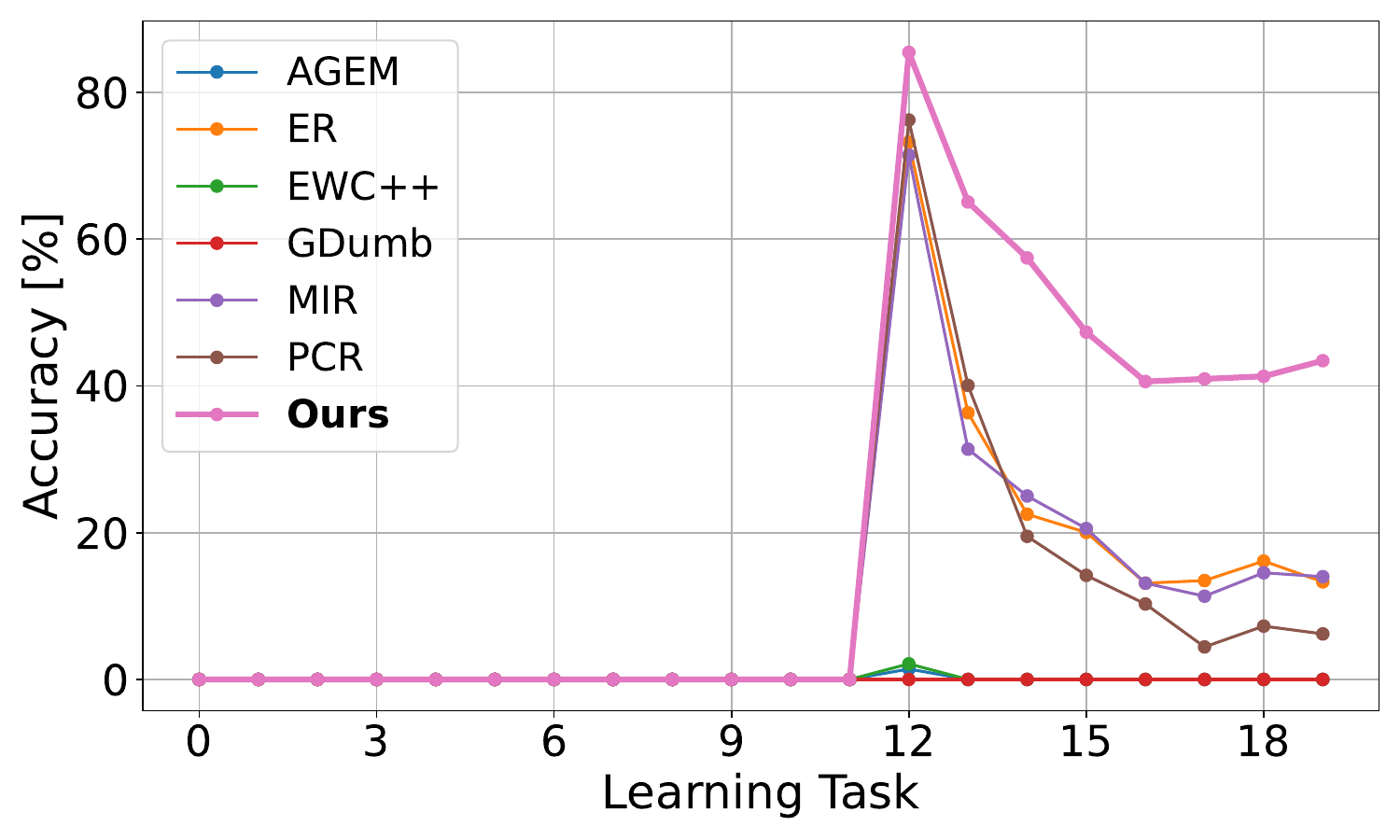}
  \caption{Task accuracy of task \#12}
  \label{fig:supp-task12}
\end{subfigure}
\begin{subfigure}{.5\textwidth}
  \centering
  \includegraphics[width=.9\linewidth]{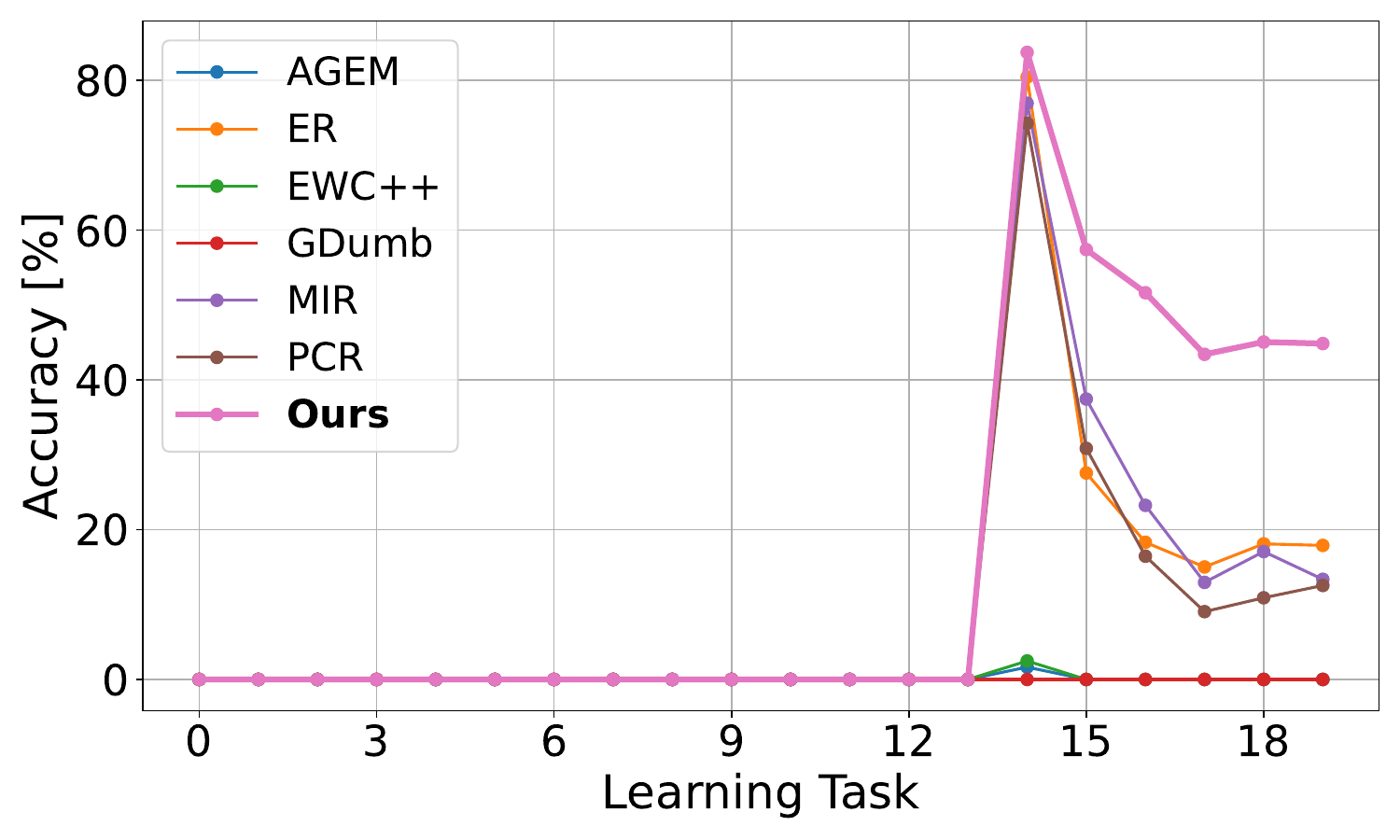}
  \caption{Task accuracy of task \#14}
  \label{fig:supp-task14}
\end{subfigure}%
\begin{subfigure}{.5\textwidth}
  \centering
  \includegraphics[width=.9\linewidth]{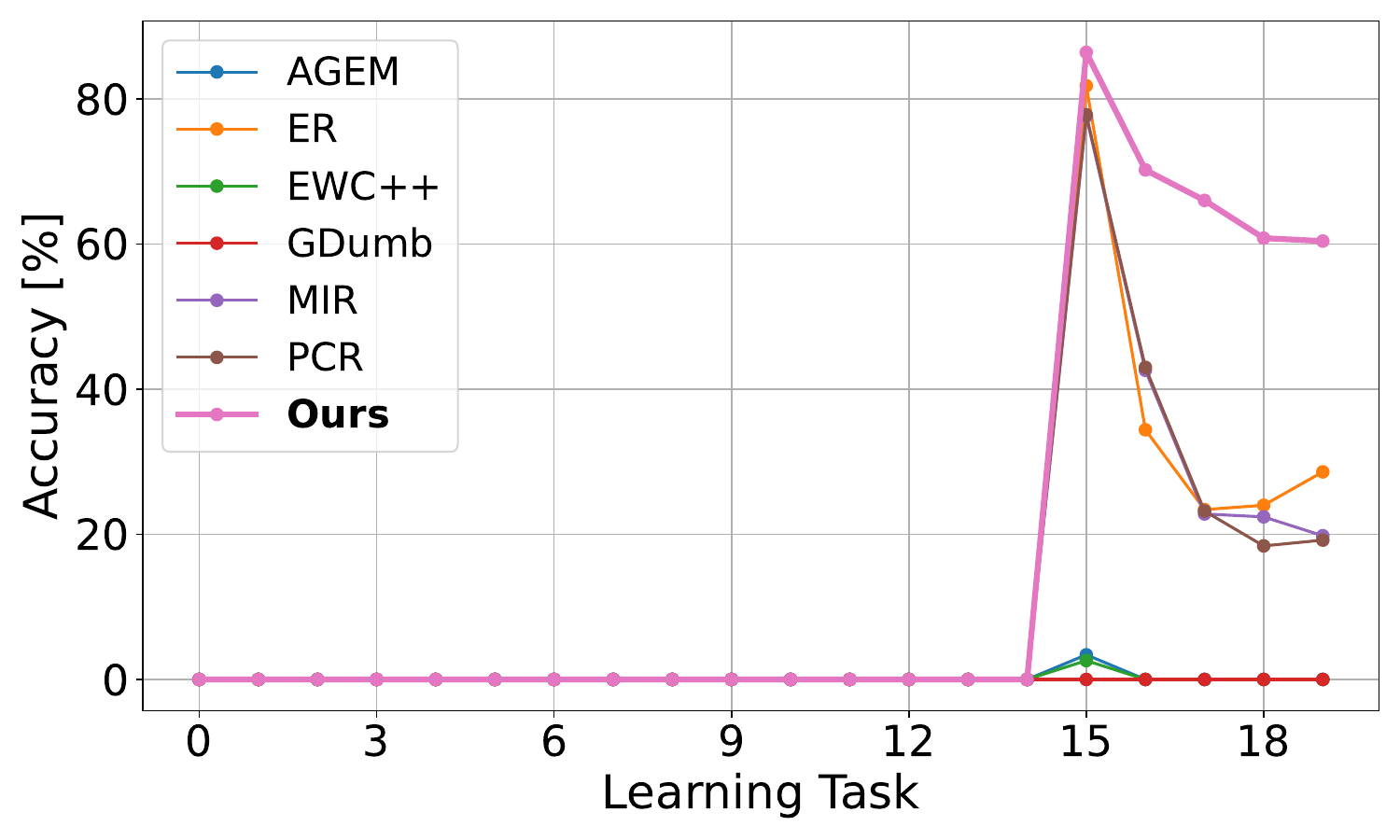}
  \caption{Task accuracy of task \#15}
  \label{fig:supp-task15}
\end{subfigure}
\begin{subfigure}{.5\textwidth}
  \centering
  \includegraphics[width=.9\linewidth]{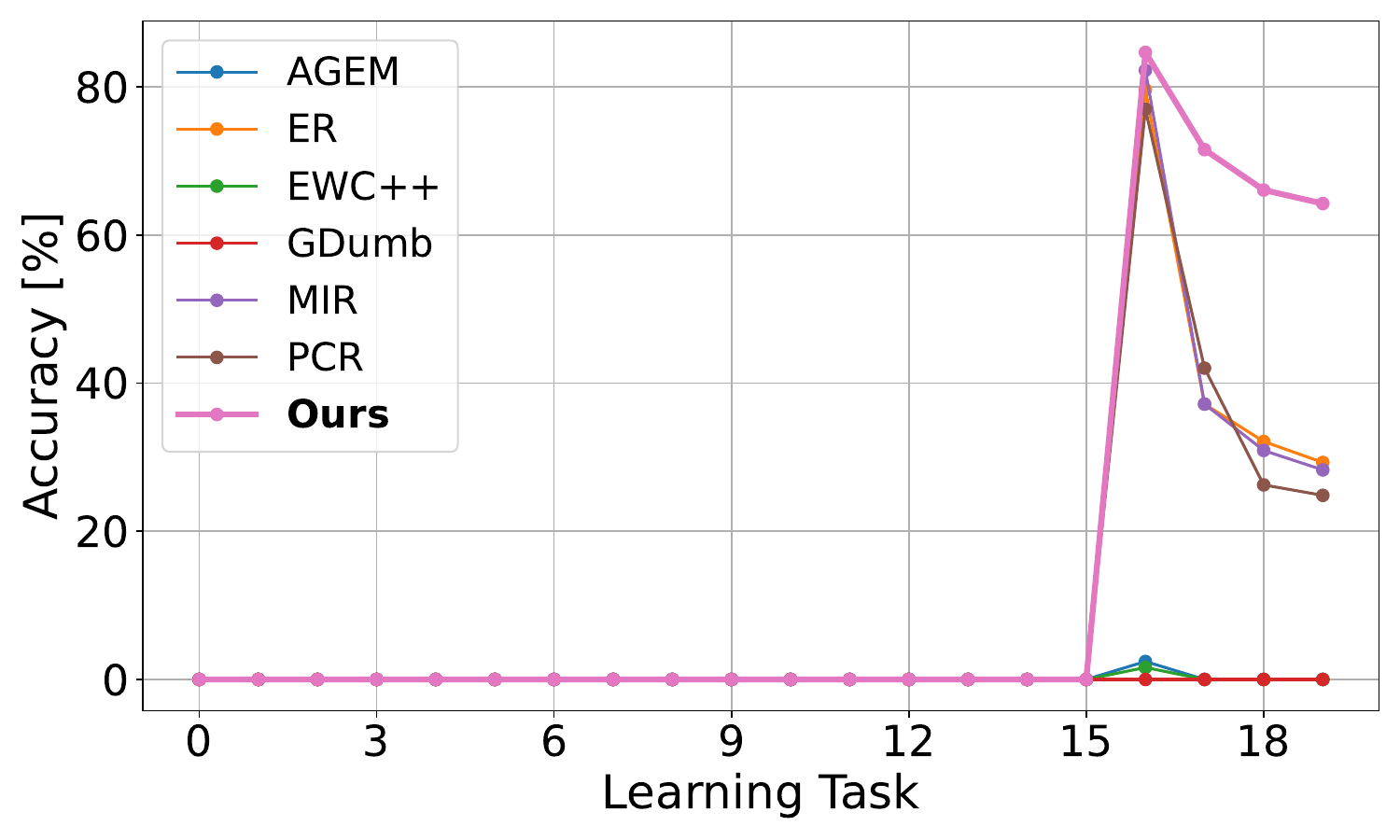}
  \caption{Task accuracy of task \#16}
  \label{fig:supp-task16}
\end{subfigure}%
\begin{subfigure}{.5\textwidth}
  \centering
  \includegraphics[width=.9\linewidth]{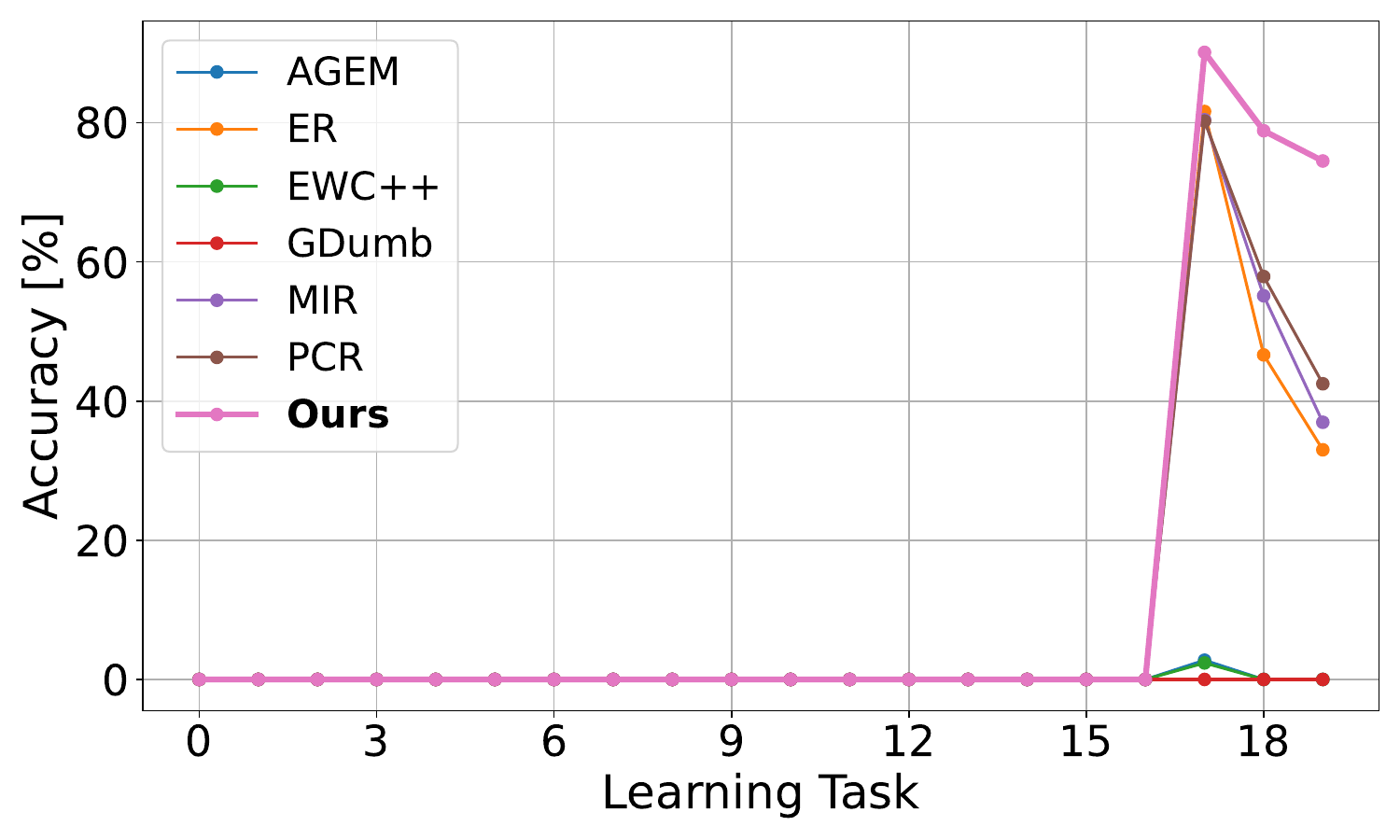}
  \caption{Task accuracy of task \#17}
  \label{fig:supp-task17}
\end{subfigure}
\caption{Task accuracy versus the number of learning tasks of task \#11 to task \#17. Compared to the results of task \#2 to task \#9 in Figure~\ref{fig:supp-task-accuracy_2-9}, our Online-LoRA has greater advantages over the other methods for these newer tasks \#11 to task \#17. Zero accuracy for initial tasks results from not measuring them at the time the specific task had not been learned yet. }
\label{fig:supp-task-accuracy_11-17}
\end{figure*}

In this section, Figure~\ref{fig:supp-task-accuracy_2-9} and Figure~\ref{fig:supp-task-accuracy_11-17} show task accuracy as a function of the number of learning tasks as described in Section~\ref{sec:main-task_length} in the main paper. The ViT-B/16 model is employed on the Split ImageNet-S dataset with 20 tasks. These results demonstrate that our Online-LoRA consistently outperforms the other methods in mitigating the forgetting of previously learned tasks. 

Figure~\ref{fig:supp-task2} shows that AGEM \cite{a-gem} begins with an initial accuracy of $\sim$10\%. However, this accuracy drastically decreases for subsequent tasks, eventually dropping to zero. Given that the Split ImageNet-S dataset consists of 20 tasks with 500 classes per task, AGEM's performance is no better than that of a random model, which would have an expected accuracy of 0.2\%. This dramatic decline is primarily due to the increasingly restrictive constraints placed on gradient updates as the number of tasks increases. Such constraints significantly hurt the model's ability to learn from new tasks, showing a fundamental weakness of AGEM in handling long sequences of diverse tasks. A similar issue was observed with EWC++ \cite{ewc++}, another regularization-based approach. 

In contrast, our Online-LoRA model does not encounter this problem even though an online weight regularization is used. This is because our model is continuously expanded by adding new LoRA parameters (see Section~\ref{main-plateau} in the main paper). This strategy allows the model to adapt to new information more flexibly, bypassing the learning limitations encountered by traditional regularization methods like AGEM and EWC++.

\section{Code}

Our code will be publicly available at: \href{https://github.com/Christina200/Online-LoRA-official.git}{https://github.com/Christina200/Online-LoRA-official.git}. Our implementation of LoRA is based on the codebase of MeLo \cite{lora-vit}. 

\clearpage
\clearpage
{\small
\bibliographystyle{ieee_fullname}
\bibliography{egbib}
}

\end{document}